\documentclass[sigconf]{acmart}
\AtBeginDocument{%
  }

\copyrightyear{2026}
\acmYear{2026}
\setcopyright{cc}
\setcctype{by}
\acmConference[WWW '26] {Proceedings of the ACM Web Conference 2026}{April 13--17, 2026}{Dubai, United Arab Emirates.}
\acmBooktitle{Proceedings of the ACM Web Conference 2026 (WWW '26), April 13--17, 2026, Dubai, United Arab Emirates}
\acmISBN{979-8-4007-2307-0/2026/04}
\acmDOI{10.1145/3774904.3792600}

\usepackage{textcomp}
\usepackage{fancyhdr}
\usepackage{tabularx}
\usepackage{ragged2e}
\newcolumntype{P}[1]{>{\Centering\hspace{0pt}}p{#1}}
\newcolumntype{Z}{>{\centering\let\newline\\\arraybackslash\hspace{0pt}}X}
\usepackage{graphicx}
\usepackage{algorithm}
\usepackage{algorithmic}
\usepackage[utf8]{inputenc} 
\usepackage{hyperref}       
\usepackage{url}            
\usepackage{booktabs}       
\usepackage{nicefrac}       
\usepackage{microtype}      
\usepackage{multirow}
\usepackage{multicol}
\usepackage{colortbl}
\usepackage{diagbox}
\newcommand{\tabincell}[2]{\begin{tabular}{@{}#1@{}}#2\end{tabular}}
\usepackage{color,xcolor}
\usepackage{amsmath,amssymb,amsfonts}
\definecolor{light-gray0}{gray}{0.85}
\usepackage{balance}

\begin{document}

\title{DyMRL: Dynamic Multispace Representation Learning for Multimodal Event Forecasting in Knowledge Graph}

\author{Feng Zhao}
\orcid{0000-0001-7205-3302}
\affiliation{%
  \department{Natural Language Processing and Knowledge Graph Lab,\\
  School of Computer Science and Technology}
  \institution{Huazhong University of Science and Technology}
  \city{Wuhan}
  \country{China}
}
\email{zhaof@hust.edu.cn}

\author{Kangzheng Liu}
\orcid{0000-0002-6362-7148}
\affiliation{%
  \department{Natural Language Processing and Knowledge Graph Lab,\\
  School of Computer Science and Technology}
  \institution{Huazhong University of Science and Technology}
  \city{Wuhan}
  \country{China}
}
\email{frankluis@hust.edu.cn}

\author{Teng Peng}
\orcid{0009-0003-2911-0473}
\affiliation{%
  \department{Natural Language Processing and Knowledge Graph Lab,\\
  School of Computer Science and Technology}
  \institution{Huazhong University of Science and Technology}
  \city{Wuhan}
  \country{China}
}
\email{pengteng@hust.edu.cn}

\author{Yu Yang}
\orcid{0000-0001-9354-3909}
\affiliation{%
  \department{Centre for Learning, Teaching and Technology}
  \institution{The Education University of Hong Kong}
  \city{Hong Kong SAR}
  \country{China}
}
\email{yangyy@eduhk.hk}

\author{Guandong Xu}
\orcid{0000-0003-4493-6663}
\affiliation{%
 \department{Centre for Learning, Teaching and Technology}
 \institution{The Education University of Hong Kong}
 \city{Hong Kong SAR}
 \country{China}
}
\email{gdxu@eduhk.hk}

\renewcommand{\shortauthors}{Feng Zhao, Kangzheng Liu, Teng Peng, Yu Yang, and Guandong Xu}

\begin{abstract}
Accurate representation of multimodal knowledge is crucial for event forecasting in real-world scenarios. However, existing studies have largely focused on static settings, overlooking the dynamic acquisition and fusion of multimodal knowledge. 1) At the knowledge acquisition level, how to learn time-sensitive information of different modalities, especially the dynamic structural modality. Existing dynamic learning methods are often limited to shallow structures across heterogeneous spaces or simple unispaces, making it difficult to capture deep relation-aware geometric features. 2) At the knowledge fusion level, how to learn evolving multimodal fusion features. Existing knowledge fusion methods based on static coattention struggle to capture the varying historical contributions of different modalities to future events. To this end, we propose \textbf{DyMRL}, a \textbf{\underline{D}}ynamic \textbf{\underline{M}}ultispace \textbf{\underline{R}}epresentation \textbf{\underline{L}}earning approach to efficiently acquire and fuse multimodal temporal knowledge. 1) For the former issue, DyMRL integrates time-specific structural features from Euclidean, hyperbolic, and complex spaces into a relational message-passing framework to learn deep representations, reflecting human intelligences in associative thinking, high-order abstracting, and logical reasoning. Pretrained models endow DyMRL with time-sensitive visual and linguistic intelligences. 2) For the latter concern, DyMRL incorporates advanced dual fusion-evolution attention mechanisms that assign dynamic learning emphases equally to different modalities at different timestamps in a symmetric manner. To evaluate DyMRL's event forecasting performance through leveraging its learned multimodal temporal knowledge in history, we construct four multimodal temporal knowledge graph benchmarks. Extensive experiments demonstrate that DyMRL outperforms state-of-the-art dynamic unimodal and static multimodal baseline methods.
\end{abstract}




\maketitle
\newcommand\webconfavailabilityurl{https://doi.org/10.5281/zenodo.18295509}
\ifdefempty{\webconfavailabilityurl}{}{
\begingroup\small\noindent\raggedright\textbf{Resource Availability:}\\
The source code of this paper has been made publicly available at \url{https://github.com/HUSTNLP-codes/DyMRL} and archived at \url{\webconfavailabilityurl}.
\endgroup
}

\section{Introduction}
\label{Introduction}
\begin{figure}[t]
\centering
\includegraphics[width=0.5\textwidth]{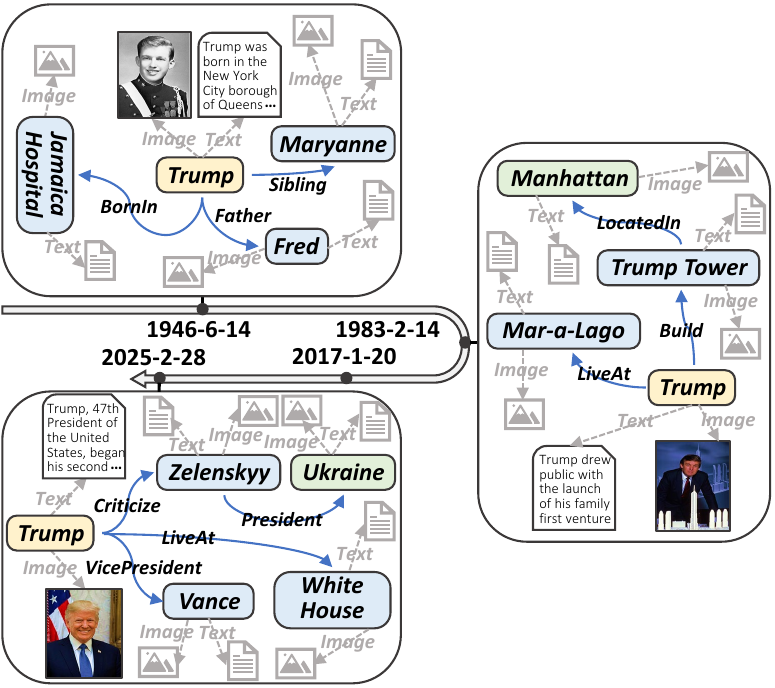}
\caption{Illustration of multimodal temporal knowledge.
}
\label{fig1}
\vspace{-1em}
\end{figure}

Multimodal knowledge graphs (KGs) find applications in diverse real-world domains, from urban management~\cite{urban} to recommendation systems~\cite{recommendation}. Effective multimodal knowledge representation is pivotal for later reasoning (e.g., event forecasting) in complex real-world scenarios. However, existing studies~\cite{DySarl,Priority} have largely focused on static settings, overlooking the dynamic acquisition and fusion of multimodal knowledge. As shown in Figure~\ref{fig1}, in the dynamic setting~\cite{DA-Net,DHU-NET}, not only structural quadruples (i.e., ($subject$, $relation$, $object$, $timestamp$)) but also rich auxiliary modality information (e.g., texts and images) evolve over time, giving rise to multimodal temporal knowledge.

According to the multi-intelligence paradigm~\cite{gardner2011frames,ryle2009concept} of human cognition, a human-like cognitive system encompasses associative thinking, high-order abstracting, logical reasoning, especially visual and language intelligences, and effectively integrate them to support future decision making. In this context, the acquisition and fusion of historical multimodal temporal knowledge provide a cognition-aligned representation learning process, which can be naturally adapted for forecasting future multimodal events.

At the knowledge acquisition level, the associative thinking, high-order abstracting, and logical reasoning intelligences derive from the dynamic structural modality (e.g., different lifelong event interactions of Trump). The visual and linguistic intelligences derive from the dynamic auxiliary modalities (e.g., different lifelong portraits and profiles of Trump). Humans collect multimodal memories for future event forecasting via diverse forms yet not a singular form of intelligence, which inspires us to integrate the inherent relational topologies in multiple spaces during dynamic structural modality learning. Moreover, as presented in~\cite{multispace1,IME}, different geometric spaces yield distinct impacts when embedding different types of structured data. For example, Euclidean space~\cite{R-GCN} models chain-like structures well, hyperbolic space~\cite{ATTH} captures hierarchical patterns, and complex space~\cite{RotatE} represents spherical shell geometries effectively. Despite many existing unimodal dynamic graph learning methods being limited to a single space~\cite{CognTKE,TempValid}, some multispace approaches based on shallow pairwise translations~\cite{IME} struggle to capture the deep graph structures among multimodal events, even when extended to multimodal settings. Hence, effectively integrating information from different geometric spaces and naturally extending it to deep structures in dynamic multimodal scenarios remains a pressing challenge.

At the knowledge fusion level, the multimodal fusion process aims to integrate the different modality features of multimodal events. However, as shown in Figure~\ref{fig1}, previous static knowledge fusion methods~\cite{MKGformer,RSME} are no longer applicable to dynamic scenarios. How to fuse dynamic features from multiple modalities in an evolving manner is an essential yet underexplored challenge. Moreover, good future event forecasting relies on effectively extracting (selecting) useful information from historical multimodal data. 
Previous static coattention-based methods~\cite{IMF,MMKGR} treat different modalities separately as attention assigners and learners, capturing only interplay between modalities without optimizing informative modality weights for later forecasting, especially in temporal scenarios.
Therefore, assigning varying emphases to different modalities at different timestamps like humans is essential for establishing fine-grained temporal dependencies between historical multimodal information and future events.

To bridge above research gaps, we propose the \textbf{DyMRL} model, which performs \textbf{\underline{Dy}}namic \textbf{\underline{M}}ultispace \textbf{\underline{R}}epresentation \textbf{\underline{L}}earning to acquire and fuse multimodal temporal knowledge for effective future multimodal event forecasting.

As shown in Figure~\ref{fig2}(a), to equip DyMRL with associative thinking intelligence, we design a Euclidean message that aggregates local neighborhood interactions directly affecting the central multimodal event via relational semantic chain-linkings. We design a hyperbolic message to replicate human high-order abstracting intelligence, enabling DyMRL to perceive global (high-order) abstract hierarchies of concurrent events through hyperbolic isometric embeddings. To equip DyMRL with logical reasoning intelligence, we design a complex message that leverage the inherent advantages of spherical shell geometry to embed four types of relational directed logics in KGs: symmetry, asymmetry, inversion, and composition. To extend above shallow geometries to deep structural modality representations, we integrate multispace messages using a carefully designed addictive attention and apply multilayer graph neural networks (GNNs) for deep message propagation. Note that the acquisition of the structural modality is then driven by update modules in parallel across $k$ historical windows. Additionally, as shown in Figure~\ref{fig2}(b), to extract dynamic visual and linguistic features, we encode and update the corresponding auxiliary modalities at each timestamp using pretrained vision and language models~\cite{bert,VGG}.

As shown in Figure~\ref{fig2}(c), to capture evolving fused features, we design a dual fusion-evolution attention mechanism, which is architecturally composed of multilayer stacks of transformer components~\cite{transformer}. To emulate human-like emphases across different modalities at different timestamps for effective future forecasting, we introduce an initialized matrix (i.e., $\textit{\textbf{E}}_{init}$) as a third-party attention assigner, with the acquired modality- and timestamp-specific embedding matrices as attention learners, respectively. Specifically, fusion attention assigns different weights to different modalities \enlargethispage{\baselineskip}
at each timestamp, while evolution attention further emphasizes different timestamps. Finally, DyMRL selectively extracts useful evolving fused features to decode and generate scores for future multimodal event forecasting.

This research makes following principal contributions.

\begin{itemize}
    \item We propose an efficient multimodal representation learning model, namely DyMRL, to fill the research gap in acquiring and fusing historical multimodal temporal knowledge for future event forecasting in dynamic scenarios.
    \item To acquire multimodal knowledge in dynamic scenarios, especially to capture dynamic deep structures with unique geometries from different spaces, we propose dynamic structural modality and auxiliary modality acquisition modules, that incorporate Euclidean, hyperbolic, and complex messages into deep message propagation, aligning with the multi-intelligence capabilities of human memory collection.
    \item To fuse multimodal knowledge in dynamic scenarios, we propose a dual fusion-evolution attention module, that dynamically assigns adaptive weights to different modalities at different timestamps, capturing temporal dependencies instead of static modality interplay between historical multimodal cues and future unknown events.
    \item We construct four multimodal temporal KG datasets to validate the effectiveness of DyMRL for multimodal event forecasting. Extensive experiments demonstrate that DyMRL significantly outperforms the state-of-the-art static multimodal and dynamic unimodal baseline methods.
\end{itemize}

In the remainder of this paper, Section~\ref{Related_Work} discusses the related work. Section~\ref{Methodology} details the DyMRL model. The experimental analyses are in Section~\ref{Experiments} before concluding in Section~\ref{Conclusion}.

\section{Related Work}
\label{Related_Work}
\subsection{Static Multimodal Forecasting Methods}
This kind of method~\cite{DySarl,IMF} aims to acquire and fuse features from different modalities (e.g., structural, visual, linguistic) in static scenarios. IKRL~\cite{IKRL} learns knowledge representations that combine images and structures, using the simple translation-based TransE~\cite{TransE} as its decoder. TransAE~\cite{TransAE} further integrates image, text, and structural information through concatenation and TransE operations. RSME~\cite{RSME} utilizes simple addition and ComplEx~\cite{ComplEx} operations to integrate visual and structural modal features. MoSE~\cite{MoSE} decouples multiple modalities and exploits ensemble inference to learn the interactions between modalities. MKGformer~\cite{MKGformer} proposes a prefix-guided attention mechanism and a similarity-aware feed forward network (FFN) to fuse the linguistic and visual modalities. OTKGE~\cite{OTKGE} employs optimal transport algorithms to compute appropriate transitions between modality representations. 
IMF~\cite{IMF} captures inter-modality correlations by combining GNNs and contrastive learning strategies. DySarl~\cite{DySarl} designs dual-space messages to learn multihop information among multimodal entities. However, static multimodal knowledge acquisition and fusion techniques are not suitable to dynamic scenarios.
\subsection{Dynamic Unimodal Forecasting Methods}
These methods~\cite{IE-Evo,FS-Net} aim to model the evolution of historical structural topologies in unimodal scenarios to effectively support future forecasting. For example, xERTE~\cite{xERTE} generates query-related inference subgraphs in history. RE-GCN~\cite{RE-GCN} models the sequential evolution process of historical event interactions over time. TiRGN~\cite{TiRGN} models periodic information in temporal sequences by encoding sine-cosine event correlations. CENET~\cite{CENET} conducts temporal contrastive learning by generating non-historical negative samples. RETIA~\cite{RETIA} and RPC~\cite{RPC} aggregate relational neighborhoods via hyperrelational GNNs. ReTIN~\cite{ReTIN} devises hyperbolic mean pooling to perceive temporal hierarchies at each historical timestamp. LogCL~\cite{LogCL} captures global long-term and local short-term historical structures. TempValid~\cite{TempValid} model historical evolutional paths based on pre-defined time-aware rules. CognTKE~\cite{CognTKE} captures the interpretable query-related forecasting paths over reconstructed cognitive directed graphs. ANEL~\cite{ANEL} designs a latent augmentation method for sparse historical knowledge. Beyond the above unispace methods, some completion methods such as IME~\cite{IME} attempt to learn multispace structures but face the following limitations: 1) pairwise translation-based paradigms capture only shallow structures; and 2) relying heavily on global information (i.e., historical, current, and future data) to complete missing events, they are incapable of forecasting future unseen events. Hence, as presented in Section~\ref{Introduction}, prior dynamic structural modality learning is limited to shallow geometries and simple unispaces. From a higher angle, unimodal dynamic methods fail to incorporate the multimodal information into the evolving process.
\begin{figure*}[t]
\centering
\includegraphics[width=0.9\textwidth]{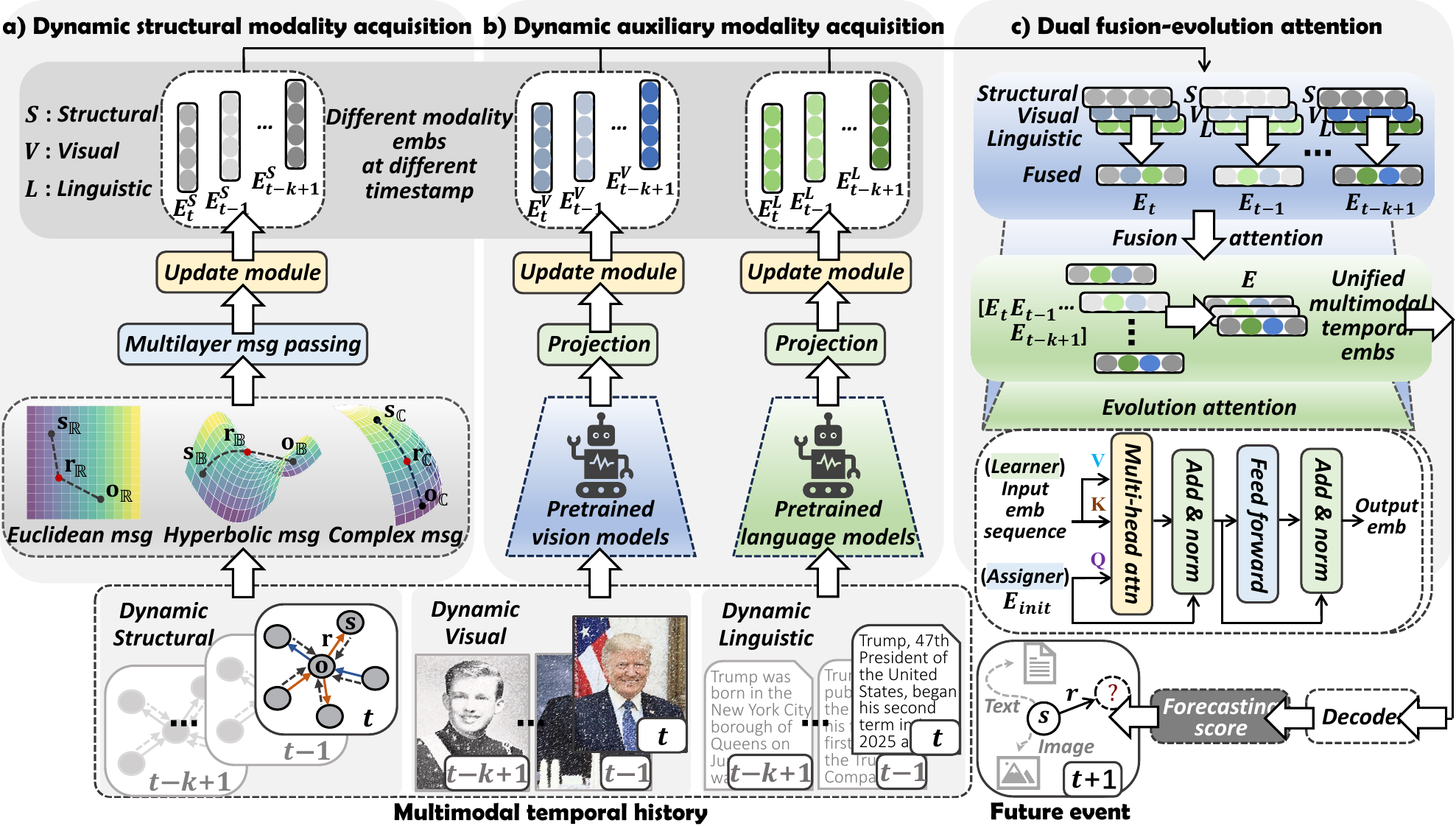}
\caption{Architecture of DyMRL model.}
\label{fig2}
\end{figure*}%

\section{The DyMRL Model}
\label{Methodology}
\subsection{Problem Definition}
We define multimodal temporal KGs and the events they contain, and formulate multimodal event forecasting as follows. Detailed notations and explanations are provided in Appendix Table~\ref{App_notations}.

\textbf{\textit{Definition 1 (Multimodal temporal KG)}}. $\mathcal{G}$ = $\{\mathcal{E},\mathcal{V},\mathcal{L},\mathcal{R},\mathcal{T},$ $\mathcal{U}\}$ extends a traditional temporal KG by augmenting each entity with time-aware visual and linguistic data alongside the dynamic structural modality. Specifically, $\mathcal{E}$, $\mathcal{V}$, $\mathcal{L}$, $\mathcal{R}$, $\mathcal{T}$, and $\mathcal{U}$ are entity, image, text, relation, timestamp, and event sets, respectively. We define the sizes of relation, entity, and timestamp sets as $R$, $N$, and $T$. $\mathcal{G}$ can be formally divided into a sequence of chronological multimodal KGs \{$\mathcal{G}_0$, $\mathcal{G}_1$, $\cdots$, $\mathcal{G}_{T-1}$\}, each of which contains all the events, images, and texts associated with a specific timestamp.

\textbf{\textit{Definition 2 (Multimodal event)}}. $\mathcal{U}$ = $\{(s,r,o,t)|s,o \in \mathcal{E},r \in \mathcal{R}, t \in \mathcal{T}\}$ indicates structural quadruples (events) in the multimodal temporal KG $\mathcal{G}$, where $o$ and $s$ denote the object and subject entities, $r$ is the relation linking them, and $t$ represents the timestamp when fact $(s,r,o)$ occurs. Note that $s$ and $o$ are augmented with auxiliary visual and linguistic modality information constrained by a specific timestamp $t$. Moreover, we define the embeddings of $s$, $r$, and $o$ in Euclidean, hyperbolic, and complex spaces as $\{\mathbf{s}_{\mathbb{R}},\mathbf{r}_{\mathbb{R}},\mathbf{o}_{\mathbb{R}}\}$, $\{\mathbf{s}_{\mathbb{B}},\mathbf{r}_{\mathbb{B}},\mathbf{o}_{\mathbb{B}}\}$, and $\{\mathbf{s}_{\mathbb{C}},\mathbf{r}_{\mathbb{C}},\mathbf{o}_{\mathbb{C}}\}$, respectively. In DyMRL, their relationships are $\mathbf{e}_{\mathbb{R}} = \mathrm{real}(\mathbf{e}_{\mathbb{C}})$ and $\mathbf{e}_{\mathbb{B}} = \exp(\mathbf{e}_{\mathbb{R}})$, where $\mathbf{e} = \{\mathbf{s}, \mathbf{r}, \mathbf{o}\}$ denotes any vector in the corresponding space. $\mathrm{real(\cdot)}$ and $\exp_{c_r}(\cdot)$ denote the real and exponential mapping operations (see Appendix Equations~(\ref{App_eq7}) and~(\ref{App_eq1})). The initial entity and relation embedding matrices $\textit{\textbf{E}}_{init}\in \mathbb{R}^{N\times d}$ and $\textit{\textbf{R}}\in \mathbb{R}^{R\times d}$ are defined in Euclidean space, where $\{\mathbf{s}_{\mathbb{R}},\mathbf{o}_{\mathbb{R}}\}\in \textit{\textbf{E}}_{init}$, $\mathbf{r}_{\mathbb{R}}\in\textit{\textbf{R}}$, and $d$ is defined as the embedding dimensionality.

\textbf{\textit{Problem 1 (Multimodal event forecasting)}}. This problem aims at forecasting missing events (i.e., a missing object $(s,r,?,t$+$1)$ or a missing subject $(?,r,o,t$+$1)$) in a future multimodal KG $\mathcal{G}_{t+1}$ when a $k$-length historical multimodal KG sequence $\{\mathcal{G}_{\tau}|t$-$k$+$1 \le \tau \textless t$+$1\}$ is known. We unify these two tasks as object forecasting by adding inverse-relation multimodal events.

\subsection{Framework Overview}
As presented in Figure~\ref{fig2}, our proposed DyMRL model is composed of three modules. Specifically, the a) dynamic structural modality acquisition module is responsible for learning the unique deep temporal topology (structure) among multimodal entities from different geometric spaces. The b) dynamic auxiliary modality acquisition module is designed to encode the auxiliary modality information (i.e., linguistic and visual) of multimodal events evolving over time. The c) dual fusion-evolution attention module aims to further equally capture the dynamic varying influence of different modalities at different timestamps on future events through two layers of carefully designed symmetric attention. Finally, the historical unified multimodal temporal embeddings are fed into a curvature-adaptive decoder to produce multimodal event forecasting scores.

\subsection{Dynamic Structural Modality Acquisition}
As shown in Figure~\ref{fig2}(a), this stage equips our devised DyMRL model with deep associative thinking, high-order abstracting, and logical reasoning intelligences by leveraging the inherent geometric properties of Euclidean, hyperbolic, and complex spaces to process historical multimodal event interactions like humans.

\subsubsection{Multispace message.}
For each historical multimodal KG in the $k$-length sequence, we aggregate the diverse geometric structural patterns from the neighborhood of each central multimodal event (entity $o$). To accommodate the heterogeneous multirelational setting of KGs, we denote the relation-specific neighborhoods of $o$ as $\mathcal{N}_{o}^{r}$. Further, we devise a Euclidean message (see the left message design in Figure~\ref{fig2}(a)) to capture chain-like associative features directly linked to $o$:
\begin{equation}
\label{eq1}
    \mathrm{\textbf{msg}}_{\mathbb{R}}^{s,r}=\mathbf{s}_{\mathbb{R}}+\mathbf{r}_{\mathbb{R}}
\end{equation}%
where $\mathrm{\textbf{msg}}_{\mathbb{R}}^{s,r}$, $\mathbf{s}_{\mathbb{R}}$, and $\mathbf{r}_{\mathbb{R}}\in \mathbb{R}^{d}$. Inspired by the superlinear property of negative curvatures~\cite{ATTH,ReTIN}, as shown in the middle message design of Figure~\ref{fig2}(a), hyperbolic embeddings can distinguish groups of events across different hyperbolic manifolds (i.e., Euclidean hierarchies) by learning relation-specific curvatures. Unlike the Euclidean message that captures local intra-neighborhood associations, the devised hyperbolic message leverages the superlinear property of learnable negative curvatures (denoted as $c_r$) to capture high-order inter-neighborhood hierarchies through a semantically-agnostic abstract manner:
\begin{equation}
    \label{eq2}
    \mathrm{\textbf{msg}}_{\mathbb{B}}^{s,r}=\mathrm{log}_{c_r}(\mathrm{H}_{r}(\mathbf{s}_{\mathbb{B}})\oplus^{c_r}\mathbf{r}_{\mathbb{B}})
\end{equation}%
where $\mathrm{\textbf{msg}}_{\mathbb{B}}^{s,r}\in \mathbb{R}^{d}$, $\mathbf{s}_{\mathbb{B}}$ and $\mathbf{r}_{\mathbb{B}}\in \mathbb{B}^{d}$. $\mathrm{log}_{c_r}(\cdot)$ maps a certain embedding from hyperbolic space to Euclidean space (see Appendix Equation~(\ref{App_eq1})). $\oplus^{c_r}$ indicates the Mobius addition in hyperbolic space (see Appendix Equation~(\ref{App_eq5})). $\mathrm{H}_{r}(\cdot)$ refers to Appendix Equation~(\ref{App_eq2}), which computes a combination of hyperbolic isometric reflection and rotation~\cite{ATTH} to preserve the inherent heterogeneous logics of relations during hierarchical learning.

Inspired by~\cite{RotatE}, to complement the fundamental relational logics in multimodal temporal KGs (i.e., symmetry /asymmetry /inversion /composition inherent in KGs), we design a complex message, as illustrated in the right message design of Figure~\ref{fig2}(a):
\begin{equation}
    \label{eq4}
    \mathrm{\textbf{msg}}_{\mathbb{C}}^{s,r}=\mathrm{real}(\mathbf{s}_{\mathbb{C}}\circ \mathbf{r}_{\mathbb{C}})
\end{equation}%
where $\mathrm{\textbf{msg}}_{\mathbb{C}}^{s,r}\in \mathbb{R}^{d}$ and $\mathbf{s}_{\mathbb{C}}$, $\mathbf{r}_{\mathbb{C}}\in \mathbb{C}^{d}$. $\circ$ denotes the complex Hadamard product (see Appendix Equation~(\ref{App_eq8})). We demonstrate that the complex message preserves above four relational logic patterns in Appendix Proof 1.

we introduce a FFN with a weight coefficient $\mathrm{W}_{s}\in \mathbb{R}^{d}$ as an alignment model for the additive attention operation to effectively integrate multispace messages:
\begin{equation}
    \label{eq5}
    \mathrm{\textbf{msg}}_{\mathbb{R},\mathbb{B},\mathbb{C}}^{s,r}=\sum_{\mathrm{space}=\{\mathbb{R},\mathbb{B},\mathbb{C}\}}f(\mathrm{W}_{s}\mathrm{\textbf{msg}}_{\mathrm{space}}^{s,r})\mathrm{\textbf{msg}}_{\mathrm{space}}^{s,r}
\end{equation}%
where $\mathrm{\textbf{msg}}_{\mathbb{R},\mathbb{B},\mathbb{C}}^{s,r}\in \mathbb{R}^{d}$.

\subsubsection{Multilayer message passing.}
We apply a multilayer GNN to extend the shallow pairwise multispace messages to deep geometric structures among multimodal events:
\begin{equation}
    \label{eq6}
    \mathbf{o}^{l+1}=\psi\left(\sum_{r\in \mathcal{R}} \sum_{s \in \mathcal{N}_{o}^{r}} \frac{1}{|\mathcal{N}_{o}^{r}|}\mathrm{W}_{r}^{l} \mathrm{\textbf{msg}}_{\mathbb{R},\mathbb{B},\mathbb{C}}^{s,r,l}+\mathrm{W}_{0}^{l} \mathbf{o}^{l}\right)
\end{equation}%
where $\mathbf{o}^{l}$ and $\mathbf{o}^{l+1}\in \mathbb{R}^{d}$ are embeddings of the aggregated multimodal event (entity $o$) in the $l^{th}$ and $(l$+$1)^{th}$ layers of the GNN. $\mathrm{\textbf{msg}}_{\mathbb{R},\mathbb{B},\mathbb{C}}^{s,r,l} \in \mathbb{R}^{d}$ is the $l^{th}$-layer multispace message. $\mathrm{W}_{0}^{l}$ and $\mathrm{W}_{r}^{l}$ denote learnable parameters for self-loop and relation-specific structural modality features. $\psi(\cdot)$ represents the rectified linear unit (ReLU) activation function.

\subsubsection{Update module (dynamic structural modality).}
Through Equation~(\ref{eq6}), we can obtain the entity embedding matrices containing unique structural modality features at each timestamp of the $k$-length historical multimodal KG sequence. We employ a recurrent neural network (RNN) as update modules to capture chronological time shifts of dynamic structural modality:
\begin{equation}
    \label{eq7}
    \textit{\textbf{E}}^{S}_{t}=\mathrm{Update}(\mathrm{DMS}_{t}(g_t,\textit{\textbf{E}}_{init}))
\end{equation}%
where $\textit{\textbf{E}}^{S}_{t}\in \mathbb{R}^{N\times d}$. $g_t$ denotes the unique graph topological information of the $t^{th}$ historical multimodal KG. $\mathrm{DMS}_{t}(\cdot)$ is the dynamic multispace structural modality acquisition procedure at the $t^{th}$ timestamp. $\mathrm{Update}(\cdot)$ is formulated as follows:
\begin{equation}
    \label{eq8}
    \textit{\textbf{E}}^{S}_{t}=\mathrm{RNN}(\textit{\textbf{E}}^{S}_{t-1},\textit{\textbf{E}}^{DMS}_{t-1})
\end{equation}%
where $\textit{\textbf{E}}^{S}_{0}$ = $\textit{\textbf{E}}_{init}$ at the $(t$-$k$+$1)^{th}$ timestamp. $\textit{\textbf{E}}^{DMS}_{t-1}\in\mathbb{R}^{N\times d}$ is the output of the $(t$-$1)^{th}$ $\mathrm{DMS}_{t-1}(\cdot)$.

\subsection{Dynamic Auxiliary Modality Acquisition}
As shown in Figure~\ref{fig2}(b), this stage equips DyMRL with visual and linguistic intelligences by adopting evolving pretrained models.

\subsubsection{Dynamic visual modality.}
We utilize pretrained VGG vision models~\cite{VGG} to acquire image features uniquely at each historical timestamp. Then, an update module is introduced to model chronological time shifts of dynamic visual modality:
\begin{equation}
    \label{eq9}
    \textit{\textbf{E}}^{V}_{t}=\mathrm{Update}(\mathrm{W}_{I}\|{pooling}(\mathrm{VGG}_t(\mathcal{V}_{t}))\|)
\end{equation}%
where $\textit{\textbf{E}}^{V}_{t}\in \mathbb{R}^{N\times d}$. $\mathcal{V}_{t} \subset \mathcal{V}$ denotes the collection of images attached to each event (entity) at the $t^{th}$ timestamp. $\mathrm{VGG}_t(\cdot)$ denotes the $t^{th}$ pretrained visual models. $pooling(\cdot)$ computes mean pooling of all attached image embeddings for each entity. Learnable coefficient $\mathrm{W}_{I}\in \mathbb{R}^{d_{v}\times d}$ projects the dimensionality of acquired image features (i.e., $d_v$) to $d$. $\mathrm{Update}(\cdot)$ refers to Equation~(\ref{eq8}). $\|\cdot\|$ indicates the L2 normalization.

\subsubsection{Dynamic linguistic modality.}
We use parallelized pretrained BERT language models~\cite{bert} to acquire time-sensitive text features. Similarly, an update module is introduced to model chronological time shifts of dynamic linguistic modality:
\begin{equation}
    \label{eq10}
    \textit{\textbf{E}}^{L}_{t}=\mathrm{Update}(\mathrm{W}_{L}\|\mathrm{BERT}_t(\mathcal{L}_{t})\|)
\end{equation}%
where $\textit{\textbf{E}}^{L}_{t}\in \mathbb{R}^{N\times d}$. $\mathcal{L}_{t}$ indicates the collection of time-sensitive textual descriptions attached to each event (entity) at the $t^{th}$ timestamp. $\mathrm{BERT}_t(\cdot)$ denotes the $t^{th}$ pretrained language models. $\mathrm{W}_{L}\in \mathbb{R}^{d_l \times d}$ projects the dimensionality of acquired text features (i.e., $d_l$) to $d$. $\mathrm{Update}(\cdot)$ refers to Equation~(\ref{eq8}).

\subsection{Dual Fusion-Evolution Attention}
As shown in Figure~\ref{fig2}(c), this stage aims to dynamically fuse multimodal features across different timestamps to build temporal dependencies between the future unseen events and historical multimodal data. Specifically, we equip DyMRL with the human-like capability to assign different emphases to different modality features at different timestamps.

\subsubsection{Fusion attention.}
This attention is designed to fuse specific multimodal features (i.e., structural, visual, and linguistic) in parallel at each timestamp within the $k$-length historical sequence. Taking the fusion process at the $t^{th}$ timestamp as an example, unlike prior coattention-based methods that focus only on modality interplay and ignore dynamic effects on future events, we use a third-party initialized matrix $\textit{\textbf{E}}_{init}$ as the attention assigner to generate the query of multi-head attention (MHA), while treating different modality matrices equally as attention learners to produce the key and value, as illustrated in the lower block of Figure~\ref{fig2}(c):
\begin{equation}
    \label{eq11}
    \textit{\textbf{E}}^{MHA}_{t}
    =\psi(\frac{\mathrm{W}_{q}\textit{\textbf{E}}_{init}(\mathrm{W}_{k}[\textit{\textbf{E}}^{S}_{t};\textit{\textbf{E}}^{V}_{t};\textit{\textbf{E}}^{L}_{t}])^\mathrm{T}}{\sqrt{d_k}})\mathrm{W}_{v}[\textit{\textbf{E}}^{S}_{t};\textit{\textbf{E}}^{V}_{t};\textit{\textbf{E}}^{L}_{t}]
\end{equation}%
where $\textit{\textbf{E}}^{MHA}_{t} \in \mathbb{R}^{N\times d}$. $[;]$ denotes the concatenation operation, and $d_k$ is a scaling factor to mitigate gradient vanishing. $\mathrm{W}_{k}$, $\mathrm{W}_{v}$, and $\mathrm{W}_{q}$ are learnable parameters for weighting different modality features at the $t^{th}$ timestamp. We introduce a FFN with $d$ hidden units to enhance temporal semantics:
\begin{equation}
    \label{eq12}
    \textit{\textbf{E}}_{t}=\mathrm{W}_{ffn1}(\psi(\mathrm{W}_{ffn2}\textit{\textbf{E}}^{MHA}_{t}))
\end{equation}%
where $\textit{\textbf{E}}_{t}\in \mathbb{R}^{N \times d}$ indicates the output embedding matrix of the $t^{th}$ fusion attention. $\mathrm{W}_{ffn1}$ and $\mathrm{W}_{ffn2} \in \mathbb{R}^{d \times d}$. Following standard practice, layer normalization and residual connections are applied after both MHA and FFN operations.
\begin{table*}[t]
\centering
\caption{Statistical information of the constructed multimodal temporal KG datasets.}
\resizebox{1.0\textwidth}{!}{
\begin{tabular}{l|cccccccccc}
\toprule
Datasets & $|\mathcal{E}|$ & $|\mathcal{V}|$ & $|\mathcal{L}|$ & $|\mathcal{R}|$ & \#Historical & \#Current & \#Future & \#Time granularity  & \#Time range & \#Timestamp \\
\midrule
GDELT-IMG-TXT   & 7,691 & 59,196 & 7,691 & 240 & 1,734,399 & 238,765 & 305,241 & 15 mins &  2018/1/1--2018/1/31  & 2,751 \\
ICE14-IMG-TXT  & 7,128 & 46,089 & 7,128 & 230 & 74,845 & 8,514 & 7,371 & 24 hours  & 
 2014/1/1--2014/12/31  & 365 \\
ICE0515-IMG-TXT  & 10,488 & 70,001 & 10,488 & 251 & 368,868 & 46,302 & 46,159 & 24 hours & 2005/1/1--2015/12/31  & 4,017 \\
ICE18-IMG-TXT  & 23,033 & 111,624 & 23,033 & 256 & 373,018 & 45,995 & 49,545 & 24 hours 
 & 2018/1/1--2018/10/31 & 304 \\
\bottomrule
\end{tabular}}
\label{tab:datasets}
\end{table*}%

\subsubsection{Evolution attention.}
This attention is devised to further assign dynamic emphases to the fused multimodal features (matrices) at different historical timestamps, enabling the extraction of time-varying informative evolving patterns that facilitate forecasting of future unknown multimodal events:
\begin{equation}
    \label{eq13}
    \textit{\textbf{E}}=\mathrm{FFN}(\mathrm{MHA}(\textit{\textbf{E}}_{init},[\textit{\textbf{E}}_{t-k+1};\cdots;\textit{\textbf{E}}_{t-1};\textit{\textbf{E}}_{t}]))
\end{equation}%
where $\textit{\textbf{E}}\in \mathbb{R}^{N \times d}$ denotes the unified multimodal temporal event (entity) embedding (matrix). $\mathrm{MHA}(\cdot)$ and $\mathrm{FFN}(\cdot)$ refer to Equation~(\ref{eq11}) and Equation~(\ref{eq12}), respectively.

\subsection{Training Strategy}
For a future multimodal query $(s,r,?,t$+$1)$, to accommodate changing curvatures during multispace learning, we adopt curvature-adaptive hyperbolic distance to convert the above unified multimodal temporal embeddings into forecasting scores:
\begin{equation}
    \label{eq14}
    \mathcal{S}=
    -\mathrm{d}^{c_r}((\mathbf{s}_{\mathbb{B}}\oplus^{c_r}\mathbf{r}_{\mathbb{B}}), \textit{\textbf{E}})^2+b_s+\textit{\textbf{b}}_{o}
\end{equation}%
where $\mathcal{S}\in \mathbb{R}^{N}$. $\mathbf{s}_{\mathbb{B}}\in \mathbb{B}^{d} \subset \exp_{c_r}(\textit{\textbf{E}})$ and $\mathbf{r}_{\mathbb{B}} \in \mathbb{B}^{d} \subset \exp_{c_r}(\textit{\textbf{R}})$.  $b_s\in \mathbb{R}$ and $\textbf{\textit{b}}_{o}\in \mathbb{R}^{N}$ are learnable biases for subject $s$ and candidate objects $o$. $\mathrm{d}^{c_r}(\cdot)$ refers to the distance computation procedure as presented in Appendix Equation~(\ref{App_eq6}).

We adopt a multilabel learning framework to train our DyMRL model through the cross-entropy loss function:
\begin{equation}
    \label{eq15}
    \mathcal{L}=\sum_{\tau=0}^{\hat{T}-1}\sum_{(s,r,o,\tau+1)\in \mathcal{G}_{\tau+1}}\sum_{i=0}^{N-1}y_{\tau+1,i}\log(\mathcal{S}_i)
\end{equation}%
where $\hat{T}$ denotes historical timestamps. $y_{\tau+1,i}$ is 1 if the historical multimodal event forecast by the $i^{th}$ entity occurs at the $(\tau$+$1)^{th}$ historical training timestamp, and 0 otherwise. $\mathcal{S}_i$ denotes the score of the $i^{th}$ entity for forecasting a historical training event.

\subsection{Computational Complexity Analysis}
The dynamic structural modality acquisition module uses a GNN-based framework to roll back $k$ historical windows (timestamps) and perform multispace message passing with a depth of $L$ layers, resulting in a time complexity of $O(kLdN)$. For the dynamic auxiliary modality acquisition module, the time complexity for encoding visual features at $k$ historical timestamps is $O(kNmh\mathcal{A}p^2c^2)$, where $\mathcal{A}$ represents the image size; $h$, $p$, and $c$ denote the depth, kernel size, and maximum number of channels of the VGG model; and $m$ represents the number of images attached to each entity. The time complexity for extracting time-sensitive linguistic features is $O(kNn^2d)$, where $n$ is the maximum description length attached to each entity. For the dual fusion-evolution attention module, the time complexity of the symmetric fusion attention layers in parallel over $k$ historical timestamps is $O(k\mathcal{M}^2d)$, where $\mathcal{M}$ represents the number of modalities. Finally, the time complexity of later symmetric evolution attention layers is $O(k^2d)$.
\begin{table*}[t]
\centering
\caption{Comparison of multimodal future forecasting performance (in percentage) against static multimodal and dynamic unimodal baseline methods across four multimodal temporal KG datasets using time-aware filtered evaluation metrics.}
\resizebox{0.9\textwidth}{!}{
\begin{tabular}{cl|ccc|ccc|ccc|ccc}
\toprule
&\multirow{2}*{Model} & \multicolumn{3}{c}{GDELT-IMG-TXT} & \multicolumn{3}{c}{ICE14-IMG-TXT} & \multicolumn{3}{c}{ICE0515-IMG-TXT} & \multicolumn{3}{c}{ICE18-IMG-TXT}\\
&& MRR & H@1  & H@10 & MRR & H@1 &  H@10 & MRR & H@1  & H@10 & MRR & H@1  & H@10\\
\midrule
\multirow{5}{*}[0em]{\rotatebox{90}{\tabincell{c}{Static\\ multimodal}}} &TransAE (\textbf{IJCNN'19}) & 13.31$^{*}$ & 4.42$^{*}$ & 29.93$^{*}$ &  24.34$^{*}$  & 11.76$^{*}$ & 49.62$^{*}$ & 26.79$^{*}$ & 13.86$^{*}$ & 52.18$^{*}$ & 19.66$^{*}$ & 9.97$^{*}$ & 39.16$^{*}$ \\
&MoSE (\textbf{EMNLP'22})  & 16.76$^{*}$  & 10.01$^{*}$ & 30.00$^{*}$ &  27.18$^{*}$  &  15.59$^{*}$  & 51.29$^{*}$ & 29.71$^{*}$ & 18.21$^{*}$ & 53.45$^{*}$ & 20.88$^{*}$ & 11.25$^{*}$ & 41.09$^{*}$ \\ 
&OTKGE (\textbf{NeurIPS'22}) & 17.01$^{*}$ & 10.09$^{*}$ & 30.50$^{*}$ & 31.36$^{*}$ & 21.03$^{*}$ & 51.57$^{*}$ & 31.36$^{*}$ & 20.36$^{*}$ & 53.42$^{*}$ & 22.51$^{*}$ & 12.70$^{*}$ & 42.68$^{*}$ \\ 
&IMF (\textbf{WWW'23}) & 17.97$^{*}$ & 10.94$^{*}$ & 31.81$^{*}$ & 33.89$^{*}$ & 24.28$^{*}$ & 52.69$^{*}$ &  34.78$^{*}$  &  24.41$^{*}$  & 55.71$^{*}$ & 26.57$^{*}$ & 16.96$^{*}$ & 45.48$^{*}$ \\
&DySarl (\textbf{MM'24}) & 30.81$^{*}$ & 18.27$^{*}$ & 46.40$^{*}$ & 38.14$^{*}$ & 21.75$^{*}$ & 57.80$^{*}$ &  47.28$^{*}$  &  29.65$^{*}$  & 69.67$^{*}$ & 33.52$^{*}$ & 21.14$^{*}$ & 52.00$^{*}$ \\
\midrule
\multirow{5}{*}[-3.5em]{\rotatebox{90}{\tabincell{c}{Dynamic\\ unimodal}}} & xERTE (\textbf{ICLR'21}) &  18.09$^{*}$ &  12.30$^{*}$ & 30.34$^{*}$ &  40.02$^{*}$  &  32.06$^{*}$  &  56.17$^{*}$ & 46.62$^{*}$  &  37.84$^{*}$  & 63.92$^{*}$ & 29.98$^{*}$ & 22.05$^{*}$ & 44.83$^{*}$ \\
& RE-GCN (\textbf{SIGIR'21}) &  19.64$^{*}$ &  12.42$^{*}$ & 33.69$^{*}$ &  40.39$^{*}$  &  30.66$^{*}$  &  59.21$^{*}$ &  48.03$^{*}$  &  37.33$^{*}$  & 68.27$^{*}$ & 30.58$^{*}$ & 21.01$^{*}$ & 48.75$^{*}$ \\
& TiRGN (\textbf{IJCAI'22}) &  21.67$^{*}$ & 13.63$^{*}$ & 37.60$^{*}$ &  44.75$^{*}$  &  34.26$^{*}$  &  65.28$^{*}$ &  50.04$^{*}$  &  39.25$^{*}$  & 70.71$^{*}$ & 33.70$^{*}$ & 23.16$^{*}$ & 54.25$^{*}$ \\
& RETIA (\textbf{ICDE'23}) &  20.29$^{*}$ & 12.70$^{*}$ & 35.13$^{*}$ &  46.28$^{*}$  &  36.02$^{*}$  & 66.37$^{*}$ &  53.62$^{*}$  &  42.57$^{*}$  & 74.11$^{*}$ & 35.92$^{*}$ & 25.23$^{*}$ & 56.58$^{*}$ \\
& CENET (\textbf{AAAI'23}) &  20.23$^{*}$ & 12.69$^{*}$ & 34.92$^{*}$ &  39.02$^{*}$  &  29.62$^{*}$  & 57.49$^{*}$ &  41.95$^{*}$  &  32.17$^{*}$  & 60.43$^{*}$ & 27.85$^{*}$ & 18.15$^{*}$ & 46.98$^{*}$ \\
&RPC (\textbf{SIGIR'23}) & 22.41$^{*}$ & 14.42$^{*}$ & 38.33$^{*}$ &  44.55$^{*}$  & 34.87$^{*}$ & 65.08$^{*}$ & 51.14$^{*}$ & 39.47$^{*}$ & 71.75$^{*}$ & 34.91$^{*}$ & 24.34$^{*}$ & 55.89$^{*}$ \\
&ReTIN (\textbf{CAAI'23})  & \underline{67.56$^{*}$}  & \underline{58.13$^{*}$} & \underline{86.07$^{*}$} &  \underline{52.43$^{*}$}  &  \underline{43.85$^{*}$}  & 68.75$^{*}$ & \underline{71.98$^{*}$} & \underline{66.87$^{*}$} & \underline{81.09$^{*}$} & \underline{52.73$^{*}$} & \underline{44.24$^{*}$} & \underline{69.26$^{*}$} \\ 
&LogCL (\textbf{ICDE'24}) & 23.75$^{*}$ & 14.64$^{*}$ & 42.33$^{*}$ & 48.87$^{*}$ & 37.76$^{*}$ & \underline{70.26$^{*}$} & 57.04$^{*}$ & 46.07$^{*}$ & 77.87$^{*}$ & 35.67$^{*}$ & 24.53$^{*}$ & 57.74$^{*}$ \\ 
&TempValid (\textbf{ACL'24}) & 21.88$^{*}$ & 14.37$^{*}$ & 37.00$^{*}$ & 45.78$^{*}$ & 35.50$^{*}$ & 65.06$^{*}$ &  50.31$^{*}$  &  39.46$^{*}$  & 70.55$^{*}$ & 33.50$^{*}$ & 23.91$^{*}$ & 52.33$^{*}$ \\
& CognTKE (\textbf{AAAI'25}) &  20.15$^{*}$ & 13.25$^{*}$ &  33.18$^{*}$ &  46.06$^{*}$  &  36.49$^{*}$  & 64.49$^{*}$ &  53.13$^{*}$  &  42.62$^{*}$  &  72.70$^{*}$ & 35.24$^{*}$ &  25.21$^{*}$ & 54.71$^{*}$ \\
& ANEL (\textbf{WWW'25}) &  22.06$^{*}$ & 13.85$^{*}$ & 38.33$^{*}$ &  44.74$^{*}$  &  34.40$^{*}$  & 64.66$^{*}$ &  50.59$^{*}$  &  39.69$^{*}$  & 71.27$^{*}$ & 34.04$^{*}$ & 23.53$^{*}$ & 54.71$^{*}$ \\
\midrule
&\cellcolor{light-gray0}\textbf{DyMRL} (\textbf{Ours}) &  \cellcolor{light-gray0}\textbf{79.34} & \cellcolor{light-gray0}\textbf{72.20} &  \cellcolor{light-gray0}\textbf{92.56} &  \cellcolor{light-gray0}\textbf{62.84}  &  \cellcolor{light-gray0}\textbf{56.00}  & \cellcolor{light-gray0}\textbf{76.03}  & \cellcolor{light-gray0}\textbf{75.83} & \cellcolor{light-gray0}\textbf{71.72} & \cellcolor{light-gray0}\textbf{83.36} & \cellcolor{light-gray0}\textbf{64.56} & \cellcolor{light-gray0}\textbf{57.29} & \cellcolor{light-gray0}\textbf{78.23} \\ 
&$\Delta$$Improve$. & 17.4\%  & 24.2\% & 7.54\% &  19.9\%  & 27.7\% & 8.21\% & 5.35\% & 7.25\% & 2.80\% & 22.4\% & 29.5\% & 13.0\% \\
\bottomrule
\end{tabular}}
\caption*{An asterisk (*) indicates that DyMRL statistically outperforms the compared baseline methods according to paired $t$-tests at a 95\% significance level. The best and second-best results are shown in bold and underlined, respectively.}
\label{tab:dynamic}
\vspace{-1.5em}
\end{table*}%

\section{Experiments}
\label{Experiments}
\subsection{Experimental Setup}
\subsubsection{Datasets reconstruction.}
Table~\ref{tab:datasets} presents the statistics of our four reconstructed multimodal temporal KG datasets: GDELT-IMG-TXT, ICE14-IMG-TXT, ICE18-IMG-TXT, and ICE0515-IMG-TXT.

Specifically, we enrich the conventional temporal KGs~\cite{TA-DistMult,RE-GCN,GDELT} with time-sensitive visual images and linguistic textual descriptions of entities. The time-sensitive structural modality and linguistic modality of the reconstructed ICE14-IMG-TXT, ICE0515-IMG-TXT, and ICE18-IMG-TXT datasets are collected from the Integrated Crisis Early Warning System~\cite{ICEWS}. They record the temporal political events that occur from 2014-01-01 to 2014-12-31, from 2005-01-01 to 2015-12-31, and from 2018-01-01 to 2018-10-31 with a time granularity of 1 day. The dynamic structural and linguistic modalities of the reconstructed GDELT dataset is extracted from the Global Database of Events, Language and Tone~\cite{GDELT}. It stores time series of social media events related to human behavior from 2018-01-01 to 2018-01-31, with a time granularity of 15 minutes.

Notably, the linguistic textual descriptions of multimodal temporal KGs explicitly contain temporal information, such as ``Former UK Minister Sunak was born in Southampton on 1980-05-12 and served as a Conservative MP since 2015-05-07...''. To supplement the dynamic visual modality, each entity first retrieves the timestamp range of its valid event spans. Then, using the entity name and its associated temporal span as search keywords, up to 10 images at different timestamps are crawled for each entity from Google Images. For fair comparisons in practice, we split the events within multimodal temporal KGs into historical, current, and future sets at ratios of 80\%/10\%/10\% in a chronological order.

\subsubsection{Evaluation metrics.}
We use the widely-adopted mean reciprocal ranking (MRR), Hits@1 (H@1), and Hits@10 (H@10) metrics to evaluate the multimodal event forecasting performance of tested methods. As mentioned in multiple previous works~\cite{TiRGN,RPC}, the static filtered setting is not suitable for dynamic scenarios. Thus, we employ the time-aware filtered setting~\cite{LogCL,CognTKE}. We report the mean results of object and subject forecasting. The detailed metric description are provided in Appendix Section~\ref{Metrics}.

\subsubsection{Baselines.}
The compared baselines are generally categorized into two groups. Specifically, the conventional static multimodal methods include TransAE~\cite{TransAE}, MoSE~\cite{MoSE}, OTKGE~\cite{OTKGE}, IMF~\cite{IMF}, and DySarl~\cite{DySarl}. Additionally, the dynamic unimodal forecasting approaches contain xERTE~\cite{xERTE}, TiRGN~\cite{TiRGN}, CENET~\cite{CENET}, RPC~\cite{RPC}, ReTIN~\cite{ReTIN}, RE-GCN~\cite{RE-GCN}, LogCL~\cite{LogCL}, TempValid~\cite{TempValid}, RETIA~\cite{RETIA}, ANEL~\cite{ANEL}, and CognTKE~\cite{CognTKE}. The descriptions are presented in Section~\ref{Related_Work}. Implementation details of our DyMRL and other baseline models are provided in Appendix Section~\ref{Implementation}.

\subsection{Multimodal Event Forecasting Results}
Except for certain static deep structural learning methods such as DySarl, existing dynamic unimodal methods generally outperform shallow static multimodal methods. This indirectly highlights the importance of modeling the evolving structural modality and underscores the value of further exploring multispace geometric topology in multimodal temporal KGs.
To the best of our knowledge, DyMRL is the first dynamic multimodal approach for multimodal event forecasting in KGs, performing significantly better than all baselines. This improvement is attributed to two factors.

First, static multimodal methods cannot model different modality information that evolves over time, particularly capture the different emphases of different dynamic fusion features for future event forecasting. Second, dynamic unimodal methods fail to integrate more auxiliary modalities beyond shallow or unispace evolutional topological modeling, especially the dynamic deep multispace structural modality. We observe that DyMRL achieves relatively slight improvements in future event forecasting performance over the state-of-the-art baselines on the ICE0515-IMG-TXT dataset. We attribute this to the large number of timestamps in the ICE0515-IMG-TXT dataset (see Table~\ref{tab:datasets}), which enables historical multimodal KG evolutional patterns to fit most future event features, thus reducing the influence of dynamic auxiliary modalities.

\subsection{Ablation Study}
\begin{figure*}[t]
\centering
\includegraphics[width=0.9\textwidth]{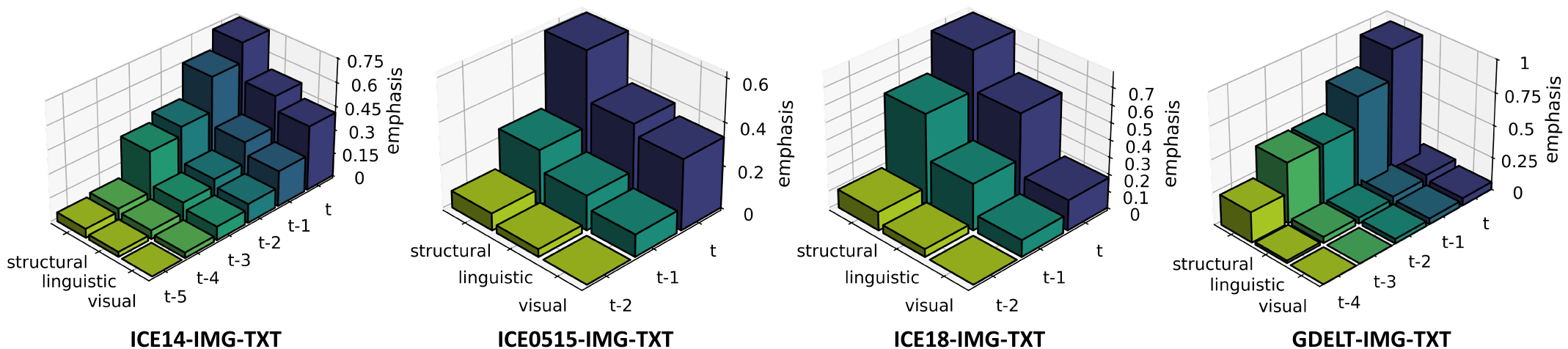}
\caption{Study on the dynamic effects of different modalities at different historical timestamps on future forecasting.}
\label{fig3}
\end{figure*}

\begin{figure}[t]
\centering
\begin{minipage}[t]{0.154\textwidth}
  \includegraphics[width=\linewidth]{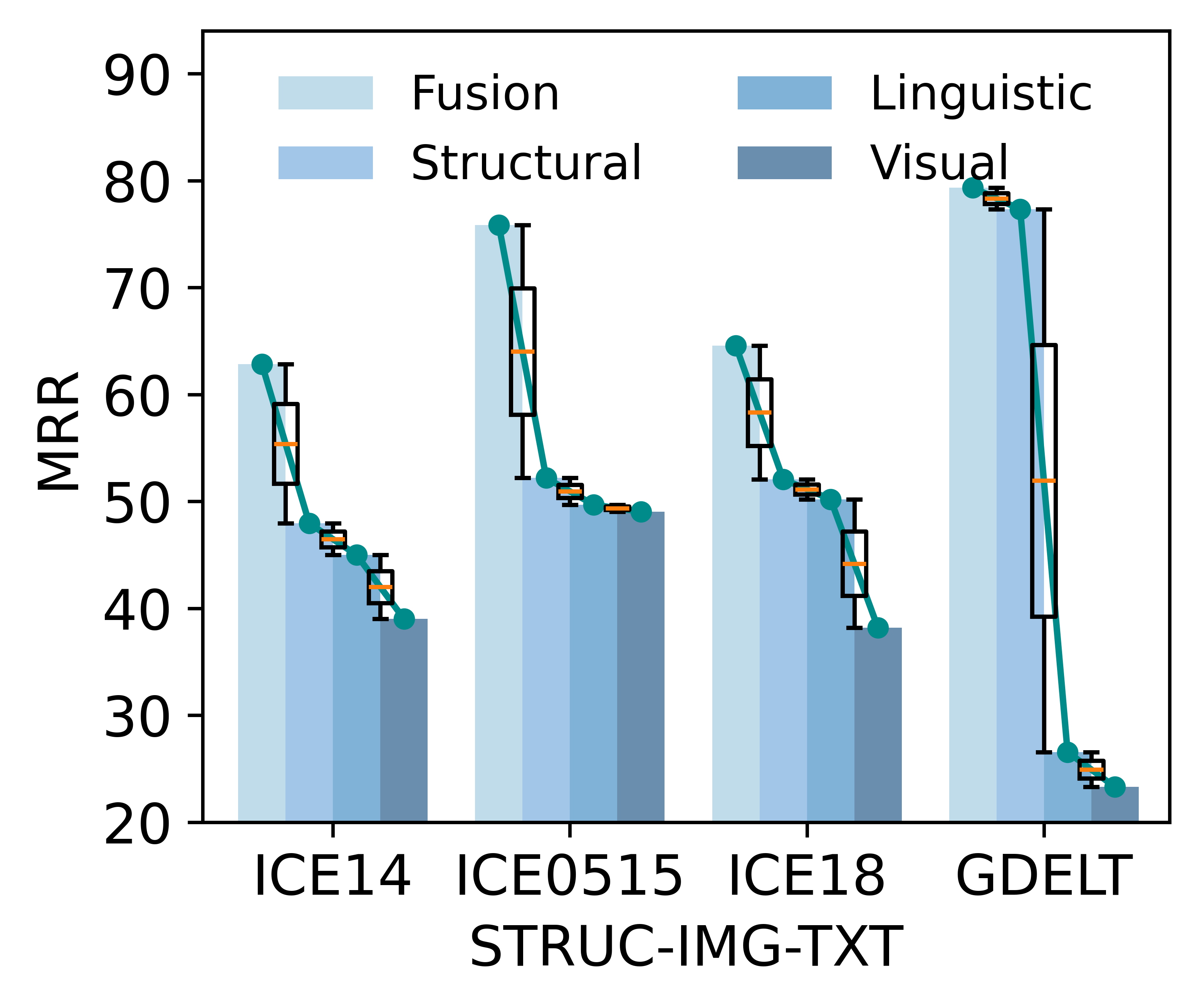}
  \caption{Study on the multiple different modalities.}
  \label{fig6}
\end{minipage}
\begin{minipage}[t]{0.154\textwidth}
  \includegraphics[width=\linewidth]{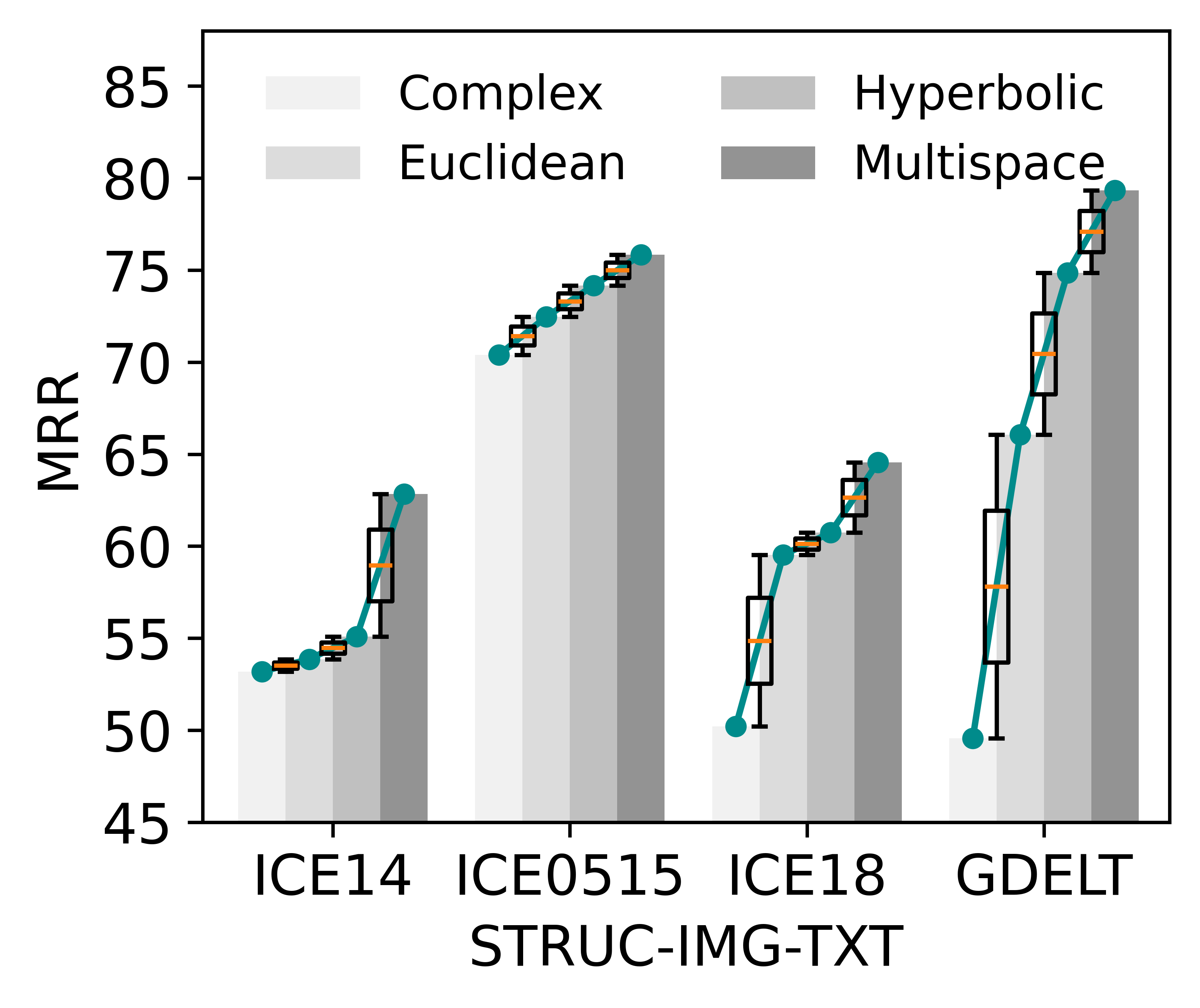}
  \caption{Study on multispace structural modality.}
  \label{fig4}
\end{minipage}
\begin{minipage}[t]{0.156\textwidth}
  \includegraphics[width=\linewidth]{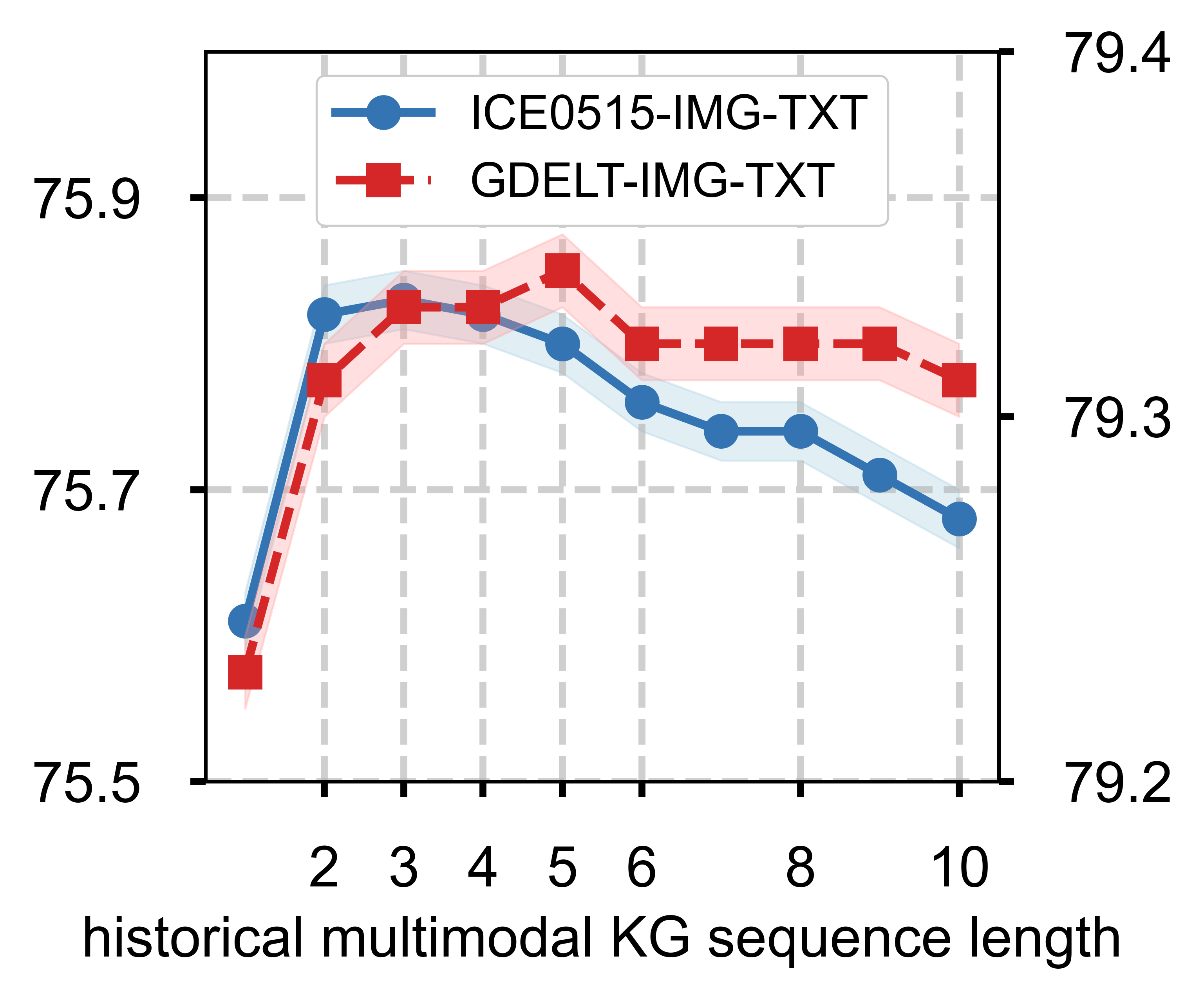}
  \caption{Study on length of historial sequence windows.}
  \label{fig7}
\end{minipage}
\vspace{-0.5em}
\end{figure}

The ablation results are presented in Table~\ref{tab:ablation}. All DyMRL variants underperform the full model, validating the effectiveness of its components, especially the geometric messages derived from different spaces. We observe a notable performance drop in DyMRL (w/o Multilayer msg propagation), as retaining only the multispace message resembles previous translation-based pairwise methods, which can capture only shallow yet not deep structural modality features. The devised update modules bring slight improvements to the forecasting performance of DyMRL. Apart from the dynamic structural modality that contributes the most, DyMRL (w/o Visual information) performs better than DyMRL (w/o Linguistic information), indicating that dynamic linguistic modality has a greater impact than the dynamic visual modality. Figure~\ref{fig6} also presents a comparison of the impact on future multimodal event forecasting between the fused modality and the individual modalities, which aligns with the above ablation results.
\begin{table}[t]
\centering
\caption{Ablation results across all the MTKG datasets.}
\resizebox{0.48\textwidth}{!}{
\begin{tabular}{l|cccc}
\toprule
Datasets (STRUC-IMG-TXT) & GDELT & ICE14 & ICE0515 & ICE18\\
\midrule
w/o Euclidean msg & 67.67 & 54.97 & 71.26 & 56.15\\
w/o Hyperbolic msg & 56.99 & 54.80 & 69.53 & 54.60\\
w/o Complex msg & 72.02 & 55.51 & 71.29 & 56.98\\
w/o Multilayer msg propagation & 31.36 & 52.98 & 62.45 & 50.51 \\
w/o Update module & 78.02 & 61.97 & 74.65 & 63.45 \\
w/o Structural information & 26.43 & 51.77 & 58.80 & 47.54\\
w/o Visual information & 72.12 & 58.36 & 69.36 & 61.49\\
w/o Linguistic information  & 70.92 & 57.30 & 62.82 & 59.99\\
\midrule
w/o Attention assigner & 45.46 & 52.06 & 44.89 & 53.77\\
w/o Fusion attention & 70.52 & 55.92 & 65.48 & 57.89\\
w/o Evolution attention & 47.47 & 53.65 & 47.35 & 56.30\\ \rowcolor{light-gray0}
\midrule
\textbf{DyMRL} & \textbf{79.34} & \textbf{62.84} & \textbf{75.83} & \textbf{64.56} \\
\bottomrule
\end{tabular}}
\vspace{-1em}
\label{tab:ablation}
\end{table}%

On the other hand, DyMRL (w/o Attention assigner) directly degenerates into prior coattention-based knowledge fusion methods~\cite{IMF,MMKGR}. Since it only captures the interplay between modalities/timestamps while overlooking their dynamic impacts on future unknown events, its performance is significantly inferior to the full DyMRL model. When the attention layers are respectively replaced with simple concatenation-projection operations, the performance of multimodal event forecasting both drops heavily, proving the effectiveness of our devised symmetric fusion-evolution attention mechanism. Furthermore, DyMRL (w/o Evolution attention) performs worse than DyMRL (w/o Fusion attention), indicating that the inter-timestamp features contribute more than inter-modality features in a historical multimodal KG sequence.

\subsection{Dynamic Multispace Structural Modality}
We investigate the impact of the unique dynamic structural features derived from different geometric spaces on the multimodal event forecasting performance of DyMRL.

As shown in Figure~\ref{fig4}, the x-axis represents different multimodal temporal KG datasets, and the y-axis shows the MRR metric results.
Among the three integrated geometries across multimodal events, the hyperbolic space provides the most effective support for future forecasting by capturing inter-neighborhood hierarchical (high-order abstracting) features. In intra-neighborhood features, chain-like (associative thinking) patterns have a greater impact than spherical (logical reasoning) patterns. This is also consistent with the ablation results shown in Table~\ref{tab:ablation}. We attribute this to the centrality of graph link structures over knowledge logic properties in large-scale KGs. Our multispace design outperforms each unispace setting, proving that DyMRL effectively captures the unique geometric features inherent in different spaces.

\subsection{Dynamic Evolving Multimodal Fusion}
As shown in Figure~\ref{fig3}, we define an attentive emphasis metric as $\beta=\frac{{MRR}_{t}^{modal}-\mathrm{min}(\tilde{MRR})}{MRR_{DyMRL}-\mathrm{min(\tilde{MRR})}}$, where $MRR_{DyMRL}$ is the MRR of full DyMRL model, ${MRR}_{t}^{modal}$ is the MRR result of a specific modality at timestamp $t$, and $\tilde{MRR}\in \mathbb{R}^{3\times k}$ stores the MRR results of all modalities across all timestamps. In Figure~\ref{fig3}, the x-axis represents different modalities, the y-axis denotes different $k$-length historical timestamps, and the z-axis indicates the emphasis ($\beta$).

It is evident from Figure~\ref{fig3} that different modalities at different timestamps contribute differently to future forecasting, and DyMRL effectively captures these dynamic multimodal evolutional fusion patterns. From the y-axis view, for any modality, short-term historical timestamps closer to the future multimodal events provide higher forecast value. From the x-axis view, at any timestamp, the structural modality provides greater forecast value than the linguistic modality, which in turn surpasses the visual modality.

\vspace{-0.2em}
\subsection{Sensitivity Study}

As shown in Figure~\ref{fig7}, the left and right y-axes represent the MRR values achieved on the ICE0515-IMG-TXT and GDELT-IMG-TXT datasets, respectively. The shaded bands around each curve denotes the MRR confidence intervals ($\pm0.02$ for ICE0515-IMG-TXT and $\pm0.01$ for GDELT-IMG-TXT) computed over 10 independent algorithm runs with different historical sequence windows. We observe that DyMRL's multimodal event forecasting performance is sensitive to the historical sequence window length $k$ (range from 1 to 10), with optimal future forecasting performance achieved at $k$ = 3 for ICE0515-IMG-TXT and $k$ = 5 for GDELT-IMG-TXT.

\section{Conclusion}
\label{Conclusion}
This paper proposes DyMRL, a dynamic multispace representation learning approach that acquires and fuses multimodal temporal knowledge from historical data in dynamic scenarios to achieve accurate future multimodal event forecasting. DyMRL integrates deep unique intrinsic geometries of Euclidean, hyperbolic, and complex spaces to acquire the dynamic structural modality. Additionally, it introduces a dual fusion-evolution attention mechanism that adaptively weights varying modalities across varying timestamps with human-like flexibility. Extensive experiments demonstrate that DyMRL significantly outperforms state-of-the-art static multimodal and dynamic unimodal forecasting baselines.

\begin{acks}
This work was supported in part by National Key Research and Development Program of China under Grant 2023YFF0905503, National Natural Science Foundation of China under Grant 62472188, EdUHK project under Grant RG 67/2024-2025R.
\end{acks}

\bibliographystyle{ACM-Reference-Format}
\balance
\bibliography{sample-base}

\appendix
\setcounter{table}{0}
\setcounter{figure}{0}
\setcounter{section}{0}
\setcounter{equation}{0}

\begin{table*}[t]
\caption{Set of notations used in the DyMRL model.}
\label{App_notations}
\begin{tabularx}{\textwidth}{P{0.08\textwidth}Z P{0.08\textwidth}Z}
\toprule
\textbf{Notation} & \textbf{Description} & \textbf{Notation} & \textbf{Description} \\
\midrule
$\mathbb{R}$ & Euclidean space & $\mathbb{B}$ & Hyperbolic space \\
$\mathbb{C}$ & Complex space & $\mathcal{G}$ & A multimodal temporal KG composed of events \\
$\mathcal{T}$ & Set of timestamps in $\mathcal{G}$ & $\mathcal{E}$ & Set of entities in $\mathcal{G}$ \\
$\mathcal{R}$ & Set of relations in $\mathcal{G}$ & $\mathcal{V}$ & Set of images in $\mathcal{G}$ \\
$\mathcal{L}$ & Set of texts in $\mathcal{G}$ & $T$ & \# of timestamps ($t \in \{0,\cdots,T{-}1\}$) \\
$N$ & \# of entities (size of $\mathcal{E}$) & $R$ & \# of relations (size of $\mathcal{R}$) \\
$\mathcal{G}_t$ & Time-specific multimodal KG valid at $t$ & $\mathbf{s}_{\mathbb{R}}$/$\mathbf{s}_{\mathbb{B}}$/$\mathbf{s}_{\mathbb{C}}$ & Euclidean/hyperbolic/complex embedding of $s$ \\
$\mathbf{o}_{\mathbb{R}}$/$\mathbf{o}_{\mathbb{B}}$/$\mathbf{o}_{\mathbb{C}}$ & Embedding of $o$ & $\mathbf{r}_{\mathbb{R}}$/$\mathbf{r}_{\mathbb{B}}$/$\mathbf{r}_{\mathbb{C}}$ & Embedding of $r$ \\
$c_r$ & Relation-specific curvature & $k$ & Historical multimodal KG sequence length \\
$d$ & Embedding dimensionality & $\textit{\textbf{E}}_{init}$ & Initial Euclidean entity embedding matrix \\
$\textit{\textbf{R}}$ & Initial Euclidean relation embedding matrix & $\textit{\textbf{E}}^{S}_{t}$ & Structural modality embedding matrix at $t$\\
$\textit{\textbf{E}}^{V}_{t}$ & Visual modality embedding matrix at $t$ & $\textit{\textbf{E}}^{L}_{t}$ & Linguistic modality embedding matrix at $t$ \\
$\textit{\textbf{E}}_{t}$ & Fused multimodal embedding matrix at $t$ & $\textit{\textbf{E}}$ & Unified multimodal temporal embeddings \\
\bottomrule
\end{tabularx}
\end{table*}%

\section{Hyperbolic Geometry}
Unlike Euclidean geometry which exhibits zero curvature, hyperbolic geometry~\cite{TKDE10,ATTH} has negative curvature, representing a form of non-Euclidean geometry. The curvature measures the extent to which a point deviates from a plane. Due to the superlinear nature of negative curvature, hyperbolic embeddings~\cite{ATTH,ReTIN} excel at capturing high-order inter-neighborhood rather than local intra-neighborhood features (i.e., tree-like geometry) in a semantically-agnostic abstract manner. 
Hence, the design of hyperbolic message that aligns high-order abstracting intelligence of humans in the dynamic structural modality acquisition stage of our proposed DyMRL extensively utilizes hyperbolic geometry. Hence, we present some basic formulas to facilitate a thorough understanding.

The exponential mapping and logarithmic mapping project a certain point (embedding) from the Euclidean space to the hyperbolic space and vice versa. They are formulated as follows:
\begin{equation}
    \begin{gathered}
    \label{App_eq1}
    \mathbf{e}_{\mathbb{B}}=\exp _{c_r}(\mathbf{e}_{\mathbb{R}})=\tanh (\sqrt{c_r}\|\mathbf{e}_{\mathbb{R}}\|) \frac{\mathbf{e}_{\mathbb{R}}}{\sqrt{c_r}\|\mathbf{e}_{\mathbb{R}}\|}\\
    \mathbf{e}_{\mathbb{R}}=\log _{c_r}(\mathbf{e}_{\mathbb{B}})=\operatorname{artanh}(\sqrt{{c_r}}\|\mathbf{e}_{\mathbb{B}}\|) \frac{\mathbf{e}_{\mathbb{B}}}{\sqrt{{c_r}}\|\mathbf{e}_{\mathbb{B}}\|}
    \end{gathered}
\end{equation}%
where $\mathbf{e}_{\mathbb{R}}\in\mathbb{R}^{\textit{d}}$ and $\mathbf{e}_{\mathbb{B}}\in\mathbb{B}^{\textit{d}}$ represent certain points (embeddings) in the Euclidean and hyperbolic spaces, respectively. $\|\cdot\|$ indicates the L2 normalization. $c_r$ denotes the relation-specific curvatures.

To preserve the inherent heterogeneous logics of relations in hierarchical learning over multimodal temporal knowledge graphs (KGs), $\mathrm{H}_{r}(\cdot)$ computes a combination of hyperbolic isometric reflection and rotation~\cite{ATTH} for a given embedding on manifolds:
\begin{equation}
\begin{aligned}
    \label{App_eq2}
    \mathrm{H}_{r}(\mathbf{e}_{\mathbb{B}})&=\mathrm{Att}(\operatorname{Ref}\left(\Phi_{r}\right)\mathbf{e}_{\mathbb{B}},\operatorname{Rot}\left(\Theta_{r}\right)\mathbf{e}_{\mathbb{B}};\boldsymbol\lambda) \\&= \exp _{c_r}(f({\boldsymbol\lambda}^\mathrm{T} \log_{c_r}(\mathbf{e}_{\mathbb{B}}^{\operatorname{Ref}}))\log_{c_r}(\mathbf{e}_{\mathbb{B}}^{\operatorname{Ref}}) \\&+f({\boldsymbol\lambda}^\mathrm{T} \log_{c_r}(\mathbf{e}_{\mathbb{B}}^{\operatorname{Rot}}))\log_{c_r}(\mathbf{e}_{\mathbb{B}}^{\operatorname{Rot}}))
    \end{aligned}
\end{equation}%
where $\mathbf{e}_{\mathbb{B}}$ and $\mathrm{H}_{r}(\mathbf{e}_{\mathbb{B}})\in \mathbb{B}^{d}$. $\exp_{c_r}(\cdot)$ and $\log_{c_r}(\cdot)$ refer to Appendix Equation~(\ref{App_eq1}). $\boldsymbol\lambda \in \mathbb{R}^{N}$ is introduced to learn the appropriate weights. $f(\cdot)$ is the Softmax activation function. $\mathbf{e}_{\mathbb{B}}^{\operatorname{Ref}}$ = $\operatorname{Ref}\left(\Phi_{r}\right)\mathbf{e}_{\mathbb{B}}$ and $\mathbf{e}_{\mathbb{B}}^{\operatorname{Rot}}$ = $\operatorname{Rot}\left(\Theta_{r}\right)\mathbf{e}_{\mathbb{B}} \in \mathbb{B}^{d}$. $\operatorname{Ref}\left(\Phi_{r}\right)$ and $\operatorname{Rot}\left(\Theta_{r}\right)$ are relation-specific block-diagonal matrices that represent the reflection and rotation operations:
\begin{equation}
\begin{aligned}
    \label{App_eq3}
    &\operatorname{Rot}\left(\Theta_{r}\right) = diag (G^{+}\left(\Theta_{r, 1}\right),\cdots,G^{+}\left(\Theta_{r, \frac{d}{2}}\right))\\
    &\operatorname{Ref}\left(\Phi_{r}\right) = diag(G^{-}\left(\Phi_{r, 1}\right),\cdots, G^{-}\left(\Phi_{r, \frac{d}{2}}\right))
    \end{aligned}
\end{equation}
where $G^{\pm}(\cdot)$ denotes the $2\times 2$ given transformation matrices:
\begin{equation}
\begin{split}
    \label{App_eq4}
G^{-}\left(\Phi_{r, i}\right)=\left[\begin{array}{cc}
\cos \Phi_{r, i} & \sin \Phi_{r, i} \\
\sin \Phi_{r, i} & -\cos \Phi_{r, i} \\
\end{array}\right]\\
G^{+}\left(\Theta_{r, i}\right)=\left[\begin{array}{cc}
\cos \Theta_{r, i} & -\sin \Theta_{r, i} \\
\sin \Theta_{r, i} & \cos \Theta_{r, i}
\end{array}\right]
\end{split}
\end{equation}%
where $\{\Phi_{r, i},\Theta_{r, i}\}_{i\in \{1,2,\cdots,\frac{d}{2}\}}$ are relation-specific learnable parameters preserving hyperbolic isometric distances.

The Mobius addition is formulated as follows:
\begin{equation}
    \label{App_eq5}
    \mathbf{e}_{\mathbb{B}} \oplus^{\textit{c}_\textit{r}} \mathbf{e}^{'}_{\mathbb{B}}=\frac{\left(1+2 {\textit{c}_\textit{r}}\langle\mathbf{e}_{\mathbb{B}}, \mathbf{e}^{'}_{\mathbb{B}}\rangle+{\textit{c}_\textit{r}}\|\mathbf{e}^{'}_{\mathbb{B}}\|^2\right) \mathbf{e}_{\mathbb{B}}+\left(1-{\textit{c}_\textit{r}}\|\mathbf{e}_{\mathbb{B}}\|^2\right) \mathbf{e}^{'}_{\mathbb{B}}}{1+2 {\textit{c}_\textit{r}}\langle\mathbf{e}_{\mathbb{B}}, \mathbf{e}^{'}_{\mathbb{B}}\rangle+{\textit{c}_\textit{r}}^2\|\mathbf{e}_{\mathbb{B}}\|^2\|\mathbf{e}^{'}_{\mathbb{B}}\|^2}
\end{equation}%
where $\mathbf{e}_{\mathbb{B}}$ and $\mathbf{e}^{'}_{\mathbb{B}}\in \mathbb{B}^{d}$ represent embeddings (points) on the hyperbolic manifolds. $\langle \cdot \rangle$ denotes the dot product operation.

Moreover, the distance between two specific points on the hyperbolic manifolds can be formulated as follows:
\begin{equation}
    \label{App_eq6}
    \mathrm{d}^{c_r}(\mathbf{e}_{\mathbb{B}}, \mathbf{e}^{'}_{\mathbb{B}})=\frac{2}{\sqrt{c_r}} \operatorname{artanh}\left(\sqrt{c_r}\left\|-\mathbf{e}_{\mathbb{B}} \oplus^{c_r} \mathbf{e}^{'}_{\mathbb{B}}\right\|\right)
\end{equation}%

\section{Complex Geometry}
Complex embeddings enhance the ability to capture relational logics by incorporating auxiliary imaginary parts for event (i.e., both entity and relation) representations. In geometry, well-trained complex embeddings~\cite{ComplEx,RotatE} form a spherical shell structure in the complex space through Hadamard product operations containing directed relational logics. We represent complex space embeddings as $\mathbf{e}_{\mathbb{C}}\in \mathbb{C}^{\textit{d}}\equiv\mathbb{R}^{2\textit{d}}$, where each complex vector is composed of a real and an imaginary part, i.e., $\mathbf{e}_{\mathbb{C}}=\mathbf{e}_{\mathbb{R}}^{\mathrm{real}}+\textit{i}\mathbf{e}_{\mathbb{R}}^{\mathrm{imag}}$. In DyMRL, we define the transformation between them as follows:
\begin{equation}
\begin{split}
    \label{App_eq7}
\mathbf{e}_{\mathbb{R}}&\triangleq \mathbf{e}_{\mathbb{R}}^{\mathrm{real}}=\mathrm{real}(\mathbf{e}_{\mathbb{C}})=\mathrm{chunk}(\mathbf{e}_{\mathbb{C}})[0],\\
&\mathbf{e}_{\mathbb{R}}^{\mathrm{imag}}=\mathrm{imag}(\mathbf{e}_{\mathbb{C}})=\mathrm{chunk}(\mathbf{e}_{\mathbb{C}})[1],
\end{split}
\end{equation}%
where $\mathrm{chunk}(\cdot)$ refers to the operation that splits the given vector into equal-length segments.

\textbf{\textit{Proof 1}}. We demonstrate that the proposed complex message, implemented as the Hadamard product between complex embeddings in complex space, effectively captures four key relational logic patterns in multimodal temporal KGs: 
\begin{itemize}
    \item \textbf{Symmetry}: If $r(s,o)$ and $r(o,s)$ hold, there exists:\\
    \begin{equation*}
    \mathbf{o}_{\mathbb{C}}=\mathbf{s}_{\mathbb{C}} \circ \mathbf{r}_{\mathbb{C}} \land \mathbf{s}_{\mathbb{C}}=\mathbf{o}_{\mathbb{C}} \circ \mathbf{r}_{\mathbb{C}} \Rightarrow \mathbf{r}_{\mathbb{C}} \circ \mathbf{r}_{\mathbb{C}} =\mathbf{1},
    \end{equation*}
    \item \textbf{Asymmetry}: If $r(s,o)$ and $\neg r(o,s)$ hold, there exists:\\
    \begin{equation*}
    \mathbf{o}_{\mathbb{C}}=\mathbf{s}_{\mathbb{C}}\circ \mathbf{r}_{\mathbb{C}} \land \mathbf{s}_{\mathbb{C}} \neq \mathbf{o}_{\mathbb{C}} \circ \mathbf{r}_{\mathbb{C}} \Rightarrow \mathbf{r}_{\mathbb{C}}\circ \mathbf{r}_{\mathbb{C}} \neq \mathbf{1},
    \end{equation*}
    \item \textbf{Inversion}: If $r_1(s,o)$ and $r_2(o,s)$ hold, there exists:\\
    \begin{equation*}
    \mathbf{o}_{\mathbb{C}}=\mathbf{s}_{\mathbb{C}} \circ \mathbf{r}_{1\mathbb{C}} \land \mathbf{s}_{\mathbb{C}}=\mathbf{o}_{\mathbb{C}}\circ \mathbf{r}_{2\mathbb{C}} \Rightarrow \mathbf{r}_{1\mathbb{C}} = \mathbf{r}_{2\mathbb{C}} ^{-1},
    \end{equation*}
    \item \textbf{Composition}: If $r_1(s,o)$, $r_2(s,o_1)$, and $r_3(o_1,o)$, there exists:\\
    \begin{equation*}
    \begin{aligned}
    &\mathbf{o}_{\mathbb{C}}=\mathbf{s}_{\mathbb{C}} \circ \mathbf{r}_{1\mathbb{C}} \land \mathbf{o}_{1\mathbb{C}}=\mathbf{s}_{\mathbb{C}} \circ \mathbf{r}_{2\mathbb{C}} \land \mathbf{o}_{\mathbb{C}}=\mathbf{o}_{1\mathbb{C}} \circ \mathbf{r}_{3\mathbb{C}}\\ &\Rightarrow \mathbf{r}_{1\mathbb{C}}=\mathbf{r}_{2\mathbb{C}} \circ \mathbf{r}_{3\mathbb{C}},
    \end{aligned}
    \end{equation*}
\end{itemize}
where $\circ$ denotes complex Hadamard product, which can be formulated as follows for $\mathbf{e}_{\mathbb{C}}$ and $\mathbf{e}^{'}_{\mathbb{C}}\in \mathbb{C}^{\textit{d}}$:
\begin{equation}
    \label{App_eq8}
    \begin{aligned}
\mathbf{e}_{\mathbb{C}}\circ \mathbf{e}^{'}_{\mathbb{C}}=&(\mathbf{e}_{\mathbb{R}}^{\mathrm{real}}*\mathbf{e}_{\mathbb{R}}^{'\mathrm{real}}-\mathbf{e}_{\mathbb{R}}^{\mathrm{imag}}*\mathbf{e}_{\mathbb{R}}^{'\mathrm{imag}})+\\ &\textit{i}(\mathbf{e}_{\mathbb{R}}^{\mathrm{real}}*\mathbf{e}_{\mathbb{R}}^{'\mathrm{imag}}+\mathbf{e}_{\mathbb{R}}^{\mathrm{imag}}*\mathbf{e}_{\mathbb{R}}^{'\mathrm{real}}),
\end{aligned}
\end{equation}%
where $*$ represents Euclidean Hadamard product (i.e., element-wise multiplication). Furthermore, $\mathbf{e}_{\mathbb{R}}^{\mathrm{real}}$, $\mathbf{e}_{\mathbb{R}}^{\mathrm{imag}}$, $\mathbf{e}_{\mathbb{R}}^{'\mathrm{real}}$, and $\mathbf{e}_{\mathbb{R}}^{'\mathrm{imag}}$ refer to Appendix Equation~(\ref{App_eq7}).

\section{Time-Aware Evaluation Metrics}
\label{Metrics}
The employed Hits@1 (H@1) and Hits@10 (H@10) indicate the proportions of events in the top 1 and 10 forecast hits, respectively, out of the total number of future multimodal query events. The mean reciprocal ranking (MRR) denotes the mean reciprocal forecasting ranking of ground-truth entities across all future events. Specifically, the abovementioned MRR and H@$C$ ($C \in \{1,10\}$)) evaluation metrics can be formulated as follows:
\begin{equation}
    \begin{gathered}
    \operatorname{MRR}=\frac{1}{\widetilde{\textit{T}}} \sum^{\widetilde{\textit{T}}}_{\tau=t_{future}} \frac{1}{|\widetilde{\mathcal{G}_{\tau}}|} \sum_{i=1}^{|\widetilde{\mathcal{G}_{\tau}}|} \frac{1}{\mathrm{rank}_i} \\
    \operatorname{H} @ \textit{C}=\frac{1}{\widetilde{\textit{T}}} \sum^{\widetilde{\textit{T}}}_{\tau=t_{future}} \frac{1}{|\widetilde{\mathcal{G}_{\tau}}|} \sum_{i=1}^{|\widetilde{\mathcal{G}_{\tau}}|} \mathbb{I}\left(\mathrm{rank}_i \leqslant \textit{C}\right)
    \end{gathered}
\end{equation}%
where $\widetilde{\textit{T}}$ represents the number of testing future timestamps and $t_{future}$ is the index of the first future timestamps. $|\widetilde{\mathcal{G}_{\tau}}|$ denotes the count of a future set containing all forecast events at a specific future timestamp (e.g., $\tau$). $\mathrm{rank}_{i}$ denotes the ground-truth entity ranking in the $i^{th}$ factual forecast results at the $\tau^{th}$ future timestamp. $\mathbb{I}(\cdot)$ is the indicator function.

In early works~\cite{CyGNet,Glean}, the static filtered setting yields better results because it filters out the global conflicting events at the current, historical, and future periods of the $(t$+$1)^{th}$ future forecasting timestamp. For example, if the event $(s,r,o_1,t$-$3)$ is valid only at the $(t$-$3)^{th}$ historical timestamp, then it is not reasonable to filter the conflicting event (entity) $o_1$ for the ground-truth answer $o$ in a future query event $(s,r,?,t$+$1)$ at the $(t$+$1)^{th}$ timestamp. However, our employed time-aware filtered setting~\cite{CognTKE,LogCL} is more reasonable because it weeds out conflicting events merely at the corresponding $(t+1)^{th}$ query timestamp for the ground-truth answer $o$ of the future event $(s,r,?,t$+$1)$.

\section{Implementation Details}
\label{Implementation}
We implement and train the DyMRL model using PyTorch on a single Tesla A100 GPU (80 GB memory) under the Ubuntu Linux operating system~\footnote{\url{https://github.com/HUSTNLP-codes/DyMRL}}. We adjust the parameters inherited from the historical set based on the model performance on the current set in terms of MRR metrics. The batch size is set as the number of multimodal events at each timestamp. We set the historical training epochs to 400 and set the patience to 100 for early stopping to guarantee model convergence and avoid overfitting. For each dataset, we conduct 10 independent algorithm runs and report the average results to reduce randomness. We use the tree-structured Parzen estimator algorithm~\cite{bergstra2015hyperopt} for heuristic search over the given candidate parameter set. Specifically, we choose the number of historical multimodal KG sequence lengths $k$ from the set $\{1,2,3,4,5,6,7,8,9,10\}$ and finally set $k$ to 6 for the ICE14-IMG-TXT dataset, 3 for the ICE0515-IMG-TXT and ICE18-IMG-TXT datasets, and 5 for the GDELT-IMG-TXT dataset. We set the embedding dimensionality $d$ to 20 for all the datasets, which is sufficient for this task.

In terms of the dynamic structural modality acquisition module, we choose the graph neural network (GNN) layers from the set $\{1,2,3,4,5\}$ and set it to $2$ for all the datasets to propagate the devised multispace messages to deep structures. For the dynamic auxiliary modality acquisition module, we generate a 4096-dimensional vector (corresponding to $d_v$) for each image across timestamps using the final fully connected layer of a pretrained VGG19 model prior to the Softmax activation. Additionally, we generate a 768-dimensional vector (corresponding to $d_l$) for each time-sensitive textual description using a pretrained BERT model. For the dual fusion-evolution attention module, the dimensionality of the key, value, and query matrices is set to 20 (i.e., attention projection matrices $\mathrm{W}_{k}$, $\mathrm{W}_{v}$, and $\mathrm{W}_{q} \in \mathbb{R}^{d \times d}$). The number of attention heads is chosen from $\{1,2,4,5,10\}$ and finally set to 2 to ensure an effective multi-head attention configuration.

In terms of the compared baselines, the time dimensions for the static multimodal methods and the auxiliary modalities for the dynamic unimodal methods are simply removed. Following the same experimental settings, baseline results for methods without open-source code, including RPC~\cite{RPC}, LogCL~\cite{LogCL}, TempValid~\cite{TempValid}, and ANEL~\cite{ANEL}, are taken from their original papers. For fair comparisons, we reproduce the results of TransAE~\cite{TransAE}, IMF~\cite{IMF}\footnote{\url{https://github.com/HestiaSky/IMF-Pytorch}}, MoSE~\cite{MoSE}\footnote{\url{https://github.com/OreOZhao/MoSE4MKGC}}, OTKGE~\cite{OTKGE}\footnote{\url{https://github.com/Lion-ZS/OTKGE}}, DySarl~\cite{DySarl}\footnote{\url{https://github.com/HUSTNLP-codes/DySarl}}, xERTE~\cite{xERTE}\footnote{\url{https://github.com/TemporalKGTeam/xERTE}}, RE-GCN\cite{RE-GCN}~\footnote{\url{https://github.com/Lee-zix/RE-GCN}}, TiRGN~\cite{TiRGN}\footnote{\url{https://github.com/Liyyy2122/TiRGN}}, CENET~\cite{CENET}\footnote{\url{https://github.com/xyjigsaw/CENET}}, RETIA~\cite{RETIA}\footnote{\url{https://github.com/CGCL-codes/RETIA}}, ReTIN~\cite{ReTIN}, and CognTKE~\cite{CognTKE}\footnote{\url{https://github.com/weichen3690/cogntke}} using their default parameter settings and the same evaluation protocols. Note that the implementations of TransAE and ReTIN are based on source codes obtained directly from the original authors.

\section{Multimodal Event Forecasting Time Comparison}
\begin{figure}[h]
\centering
\includegraphics[width=0.5\textwidth]{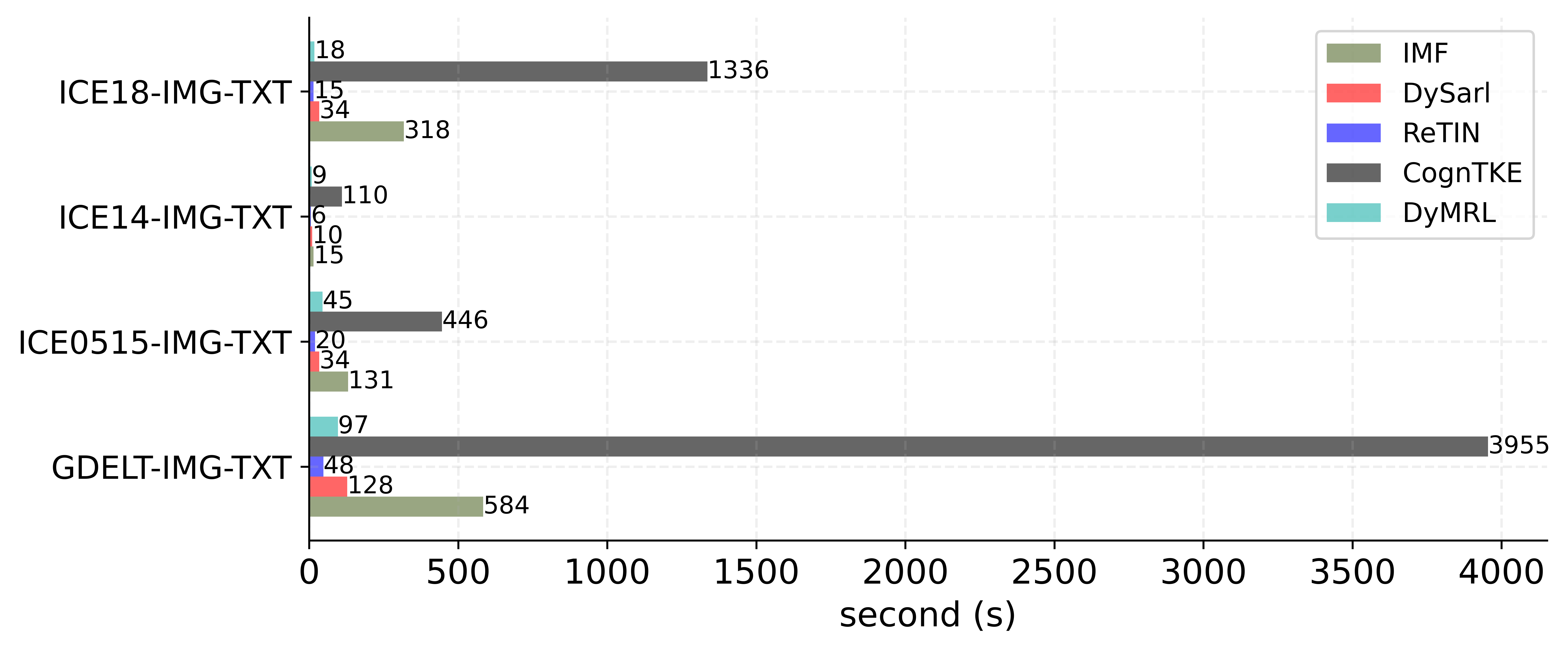}
\vspace{-2em}
\caption{Study on the multimodal event forecasting time.}
\label{App_fig1}
\end{figure}%
We investigate the time efficiency of DyMRL at the data level. As presented in Figure~\ref{App_fig1}, we plot the multimodal event forecasting time consumption of various methods on different multimodal temporal KG datasets. Specifically, the comparison baseline models include state-of-the-art static multimodal IMF~\cite{IMF}, DySarl~\cite{DySarl}, and dynamic unimodal ReTIN~\cite{ReTIN}, CognTKE~\cite{CognTKE}. It is observed that DyMRL is 40, 11, 9, and 73 times faster than CognTKE on the GDELT-IMG-TXT, ICE14-IMG-TXT, ICE0515-IMG-TXT, and ICE18-IMG-TXT datasets, respectively. This is because rule-based CognTKE processes queries one by one at each timestamp in an event-oriented manner. In contrast, Cog-RMH learns dynamic representations to obtain event embeddings for all queries simultaneously at each timestamp in a graph-oriented manner. Owing to the same reason of its parallel-friendly design, DyMRL is even faster than the static event-oriented IMF by times of 5, 0.7, 2, and 17.

Except for the GDELT-IMG-TXT, ICE14-IMG-TXT, and ICE18-IMG-TXT datasets, our DyMRL is 11 seconds slower than the static DySarl on the ICE0515-IMG-TXT dataset. This is mainly due to ICE0515-IMG-TXT having the largest number of timestamps (see Paper Table~\ref{tab:datasets}), which increases the overhead of evolutional modeling compared to DySarl, which does not model temporal dynamics. DyMRL is slower than the unimodal ReTIN by 49, 3, 25, and 3 seconds, respectively. We attribute this to the higher model complexity of DyMRL, which involves deep multispace structural modality acquisition, dynamic auxiliary modality acquisition, and evolving multimodal fusion. In summary, DyMRL demonstrates highly competitive time efficiency.

\end{document}